\documentclass{article}

\usepackage{PRIMEarxiv}

\usepackage[utf8]{inputenc} 
\usepackage[T1]{fontenc}    
\usepackage{hyperref}       
\usepackage{url}            
\usepackage{booktabs}       
\usepackage{amsfonts}       
\usepackage{nicefrac}       
\usepackage{microtype}      
\usepackage{lipsum}
\usepackage{fancyhdr}       
\usepackage{graphicx}       
\graphicspath{{images/}}     
\usepackage{amsmath}
\usepackage{subcaption}

\title{A Unified Benchmark of Deep Learning Models for Multi-task 3D Brain Tumor Segmentation from Magnetic Resonance Imaging}

\author{
  Diego J. Torrejón \\ 
  Centro de Tecnologías de la Imagen (CTIM)\\
  Instituto Universitario de Cibernética, Empresas y Sociedad (IUCES)\\
  University of Las Palmas de Gran Canaria \\
  35017 Las Palmas de Gran Canaria, Spain\\
  \texttt{diego.torrejon101@alu.ulpgc.es}\\
  \And Luna Y. Hernández \\
  Centro de Tecnologías de la Imagen (CTIM)\\
  Instituto Universitario de Cibernética, Empresas y Sociedad (IUCES)\\
  University of Las Palmas de Gran Canaria \\
  35017 Las Palmas de Gran Canaria, Spain\\
  \texttt{luna.hernandez102@alu.ulpgc.es@ulpgc.es} \\
  \And Javier Sánchez \\
  Centro de Tecnologías de la Imagen (CTIM)\\
  Instituto Universitario de Cibernética, Empresas y Sociedad (IUCES)\\
  University of Las Palmas de Gran Canaria \\
  35017 Las Palmas de Gran Canaria, Spain\\
  \texttt{jsanchez@ulpgc.es} \\
}

\begin{document}
\maketitle

\begin{abstract}
Automatic brain tumor segmentation from magnetic resonance imaging (MRI) has become a fundamental task in computer-assisted diagnosis, treatment planning, and disease monitoring. Although numerous deep learning architectures have recently been proposed, objective comparisons remain challenging because published studies often employ different datasets, preprocessing strategies, training protocols, and evaluation procedures. This work presents a unified experimental benchmark for comparing representative convolutional neural networks (CNNs), Transformer-based models, and recent State Space Model (SSM) architectures under homogeneous experimental conditions. Five state-of-the-art three-dimensional segmentation models, including 3D U-Net, SegResNet, Swin UNETR, SegMamba, and SegMambaV2, are evaluated on two brain tumor segmentation datasets representing distinct clinical scenarios: intracranial meningioma segmentation (BraTS 2023) and post-treatment glioma segmentation (BraTS 2024). All architectures are trained using identical preprocessing, data augmentation, optimization strategies, and evaluation protocols to ensure a fair comparison. Performance is assessed using segmentation accuracy metrics together with computational cost indicators, including inference time and the size of each model. The results provide practical insights into the trade-offs between segmentation accuracy and computational efficiency, highlighting the suitability of different architectural paradigms for challenging three-dimensional brain tumor segmentation tasks.
\end{abstract}


\section{Introduction}
Brain tumor segmentation from magnetic resonance imaging (MRI) constitutes one of the most important applications of artificial intelligence in medical image analysis. Accurate delineation of tumor regions provides valuable information for diagnosis, treatment planning, surgical guidance, radiotherapy, and longitudinal patient follow-up. However, manual segmentation is a labor-intensive process that requires highly specialized clinical expertise and often suffers from inter- and intra-observer variability. Consequently, the development of reliable automatic segmentation methods has become an active research topic over the last decade.

Recent advances in deep learning have substantially improved the performance of automatic medical image segmentation. Convolutional Neural Networks (CNNs), particularly U-Net~\cite{ronneberger2015u}  and its three-dimensional variants~\cite{cciccek20163d,hatamizadeh2022unetr}, established the foundations of modern medical image segmentation by effectively combining hierarchical feature extraction with multi-scale contextual information. Their success motivated numerous extensions incorporating residual connections, attention mechanisms, and more efficient encoder-decoder designs.

More recently, Transformer-based architectures have demonstrated remarkable performance by modeling long-range spatial dependencies that conventional convolutional operations cannot easily capture. Among these approaches, Swin UNETR~\cite{Swin_UNETR} combines hierarchical vision transformers with a U-shaped decoding architecture, achieving state-of-the-art results in several medical image segmentation benchmarks. Nevertheless, the quadratic computational complexity associated with self-attention has motivated the search for more computationally efficient alternatives.

State-space models (SSMs)~\cite{gu2022efficiently}, particularly architectures derived from the Mamba framework~\cite{gu2024mamba, ma2024u, ruan2024vm}, have recently emerged as a promising alternative for dense prediction tasks. These models aim to capture long-range dependencies while significantly reducing computational complexity compared to Transformer-based approaches. Architectures such as SegMamba~\cite{xing2024segmamba,xing2025segmamba} have reported competitive segmentation performance while maintaining favorable computational efficiency, making them attractive candidates for large three-dimensional medical imaging applications.

Despite the rapid evolution of segmentation architectures, fairly comparing different methods remains difficult. Published studies frequently employ different datasets, preprocessing pipelines, data augmentation strategies, optimization settings, hardware configurations, and evaluation protocols. As a consequence, reported performance differences often reflect experimental variations rather than intrinsic architectural capabilities. Moreover, many comparisons focus exclusively on segmentation accuracy while overlooking computational requirements, which are important for real-world clinical deployment.

This work addresses these limitations by developing a unified experimental benchmark to evaluate representative CNN-, Transformer-, and SSM-based segmentation architectures under identical conditions. Five representative models---3D U-Net~\cite{cciccek20163d}, SegResNet~\cite{SegResNet}, Swin UNETR~\cite{Swin_UNETR}, SegMamba~\cite{xing2024segmamba}, and SegMambaV2~\cite{xing2025segmamba}---are trained and evaluated using the same preprocessing pipeline, optimization strategy, validation protocol, and inference procedure. The evaluation is performed on two complementary public datasets from the Brain Tumor Segmentation (BraTS) challenges~\cite{Menze2015} that represent substantially different clinical scenarios: intracranial meningioma segmentation from BraTS 2023~\cite{labella2023asnrmiccai} and post-treatment glioma segmentation from BraTS 2024~\cite{BraTS24}. These datasets differ considerably in tumor morphology, anatomical complexity, and segmentation difficulty, allowing the robustness of each architecture to be analyzed under diverse conditions. 

Beyond conventional segmentation metrics, this study also evaluates computational efficiency through inference time, training cost, and the number of trainable parameters, providing a more comprehensive assessment of practical applicability.

The main contributions of this work are summarized as follows: i) We propose a unified and fully reproducible experimental framework for evaluating modern three-dimensional brain tumor segmentation architectures under identical training and evaluation conditions; ii) our study comprises a comprehensive comparison of representative CNN-, Transformer-, and State Space Model-based architectures across two clinically distinct BraTS benchmark datasets; iii) it includes a joint evaluation of segmentation accuracy and computational efficiency, providing practical insights into the trade-offs between predictive performance and computational cost; and iV) we analyze the architectural behavior changes across different brain tumor segmentation scenarios, highlighting the strengths and limitations of each family of models.

Section~\ref{sec:related_work} reviews the state of the art in deep learning-based brain tumor segmentation and recent advances in CNN-, Transformer-, and SSM-based architectures. Section~\ref{sec:materials_methods} presents the datasets, preprocessing pipeline, evaluated neural network architectures, experimental protocol, and evaluation metrics. Section~\ref{sec:results} reports the quantitative and qualitative experimental results, analyzing the strengths and limitations of the evaluated approaches, and examining the influence of the clinical scenario on model performance. Finally, Section~\ref{sec:conclusion} summarizes the main conclusions and future research directions.

\section{Related Work}
\label{sec:related_work}
Brain tumor segmentation from MRI has been one of the most active research topics in medical image analysis over the last decade~\cite{van2021performance, diana2025review, Dorfner2025, Missaoui2025}. The availability of large annotated datasets and the rapid development of deep learning techniques have considerably improved segmentation accuracy, enabling increasingly reliable computer-assisted diagnosis and treatment planning. Among the initiatives that have driven this progress, the BraTS challenge~\cite{Menze2015} has become the \textit{de facto} benchmark for evaluating segmentation algorithms under standardized conditions, fostering continuous advances.

\subsection{Convolutional Neural Networks for Brain Tumor Segmentation}
The introduction of Fully Convolutional Networks (FCNs)~\cite{FCNs15} represented a major milestone in semantic segmentation by enabling end-to-end dense prediction without requiring handcrafted features. Shortly afterward, U-Net~\cite{ronneberger2015u} established the encoder–decoder architecture that has become the foundation of most modern medical image segmentation methods. Its symmetric design, combined with skip connections between encoder and decoder stages, allows the recovery of fine spatial details while preserving high-level contextual information. The extension of U-Net to volumetric data~\cite{cciccek20163d, hatamizadeh2022unetr, Swin_UNETR} through 3D convolutions further improved segmentation performance on MRI volumes by exploiting spatial correlations across slices. 

Since then, numerous CNN-based variants have been proposed, incorporating residual learning, dense connectivity, attention mechanisms, and multi-scale feature fusion to improve gradient propagation and feature representation. Attention U-Net \cite{Oktay2018} introduced learnable attention gates that suppress irrelevant background responses while highlighting salient features passed through skip connections, a mechanism later adopted by several brain-tumor-specific architectures. Among CNN-based architectures, SegResNet \cite{SegResNet} has demonstrated strong performance while maintaining relatively low computational complexity, making it one of the most widely adopted baseline models in medical image segmentation. More broadly, nnU-Net~\cite{Isensee2021} showed that a carefully self-configured, dataset-adaptive U-Net pipeline can match or surpass purpose-built architectures across dozens of biomedical segmentation tasks, underscoring the extent to which preprocessing and training-protocol choices, not only network design, determine final performance -- a consideration directly relevant to the approach adopted in this work.

Although convolutional architectures excel at extracting local image features~\cite{reyes2024performance,medina2023high}, their inherently limited receptive field restricts their ability to model long-range spatial dependencies. While deeper networks partially alleviate this limitation, they generally increase computational cost and optimization difficulty.

\subsection{Transformer-Based Medical Image Segmentation}

Inspired by the success of Transformers~\cite{vaswani2017attention} in natural language processing and computer vision, several medical image segmentation models have incorporated self-attention mechanisms to capture global contextual information. Vision Transformers (ViTs)~\cite{dosovitskiy2020image} divide images into patches that are processed through self-attention layers, enabling direct modeling of long-range relationships between distant anatomical structures. TransUNet~\cite{chen2024transunet} was among the first hybrid designs to combine a Transformer encoder operating on CNN feature maps with a convolutional decoder, improving multi-organ and cardiac segmentation over purely convolutional baselines. In the specific context of brain tumors, TransBTS~\cite{Wang2021} adopted a similar hybrid encoder-decoder philosophy directly in 3D, using a convolutional stem for local feature extraction followed by a Transformer bottleneck for global context modeling; and UNETR \cite{hatamizadeh2022unetr} proposed a pure ViT encoder directly connected to a convolutional decoder via skip connections at multiple resolutions.

Among the architectures specifically designed for volumetric medical images, Swin UNETR~\cite{Swin_UNETR} has become one of the most influential models. Instead of applying global attention to the entire image, it employs shifted-window self-attention~\cite{liu2021swin, liu2022swin} to reduce computational complexity while maintaining hierarchical feature extraction. The transformer-based encoder is coupled with a convolutional decoder following the classical U-shaped design, combining global contextual information with precise spatial localization.

Transformer-based approaches have achieved state-of-the-art performance in numerous BraTS editions and other medical image segmentation benchmarks. They have also demonstrated effectiveness in multimodal and multi-organ classification tasks~\cite{martin2025evaluation}. However, their computational requirements remain considerably higher than those of conventional CNNs, particularly when processing high-resolution three-dimensional volumes. The quadratic complexity of the self-attention operation also limits scalability for large medical images.

\subsection{State-Space Models for Medical Image Segmentation}

Recently, SSMs~\cite{gu2022efficiently} have emerged as an alternative paradigm for sequence modeling that seeks to preserve the ability to capture long-range dependencies while reducing computational complexity. The Mamba architecture~\cite{gu2024mamba} introduced an efficient selective state-space mechanism capable of modeling long sequences with linear computational complexity, attracting considerable attention across multiple computer vision tasks. Vision Mamba \cite{Zhu2024} and VMamba \cite{Liu2024vmamba} were among the first works to adapt selective scanning to two-dimensional visual data, introducing bidirectional and cross-scan mechanisms, respectively, to compensate for the lack of a natural sequential ordering in images.

In the medical imaging domain, this line of work rapidly produced a family of Mamba-based U-shaped segmentation networks. U-Mamba \cite{ma2024u} was among the first to hybridize convolutional and Mamba blocks within a U-Net backbone, while VM-UNet \cite{ruan2024vm, Zhang2024vmunetv2}, and Mamba-UNet \cite{Wang2024mambaunet} explored purely Mamba-based or progressively refined encoder designs. Swin-UMamba \cite{Liu2024swinumambas} combined ImageNet-based pretraining with Mamba blocks to improve convergence on limited medical datasets. LKM-UNet \cite{Wang2024lkmunet} introduced large-kernel convolutional priors alongside Mamba layers to strengthen local feature extraction, and Rotate-to-Scan \cite{Tang2024} proposed a triplet state-space module with multiple scanning orientations to better capture volumetric context, a design principle conceptually related to the tri-oriented scanning strategy later adopted by SegMamba. Several recent surveys \cite{Heidari2024,Bansal2024,Rahman2024} chart the rapid expansion of this design space, highlighting both the promise and the current heterogeneity of Mamba-based architectures for medical image analysis.

Building upon 3D volumetric segmentation, SegMamba~\cite{xing2024segmamba} adapts selective state-space modeling to 3D medical image segmentation. Rather than relying on self-attention, the architecture models spatial dependencies using state-space representations integrated into a hierarchical encoder-decoder network. This design enables efficient global context modeling while reducing memory consumption compared with Transformer-based architectures. More recently, SegMambaV2~\cite{xing2025segmamba} introduced architectural refinements to improve feature extraction and representation learning while preserving the computational advantages of state-space models. Although both architectures have reported competitive performance on several public benchmarks, their evaluation has generally been limited to individual datasets and heterogeneous experimental settings, making direct comparison with CNN- and Transformer-based models difficult.

\subsection{Benchmark Datasets for Brain Tumor Segmentation}

The BraTS challenges~\cite{Menze2015} have played a fundamental role in advancing automatic brain tumor segmentation by providing publicly available datasets, standardized evaluation metrics, and common benchmarking protocols. Successive editions have progressively addressed increasingly challenging clinical scenarios. The original benchmark, later extended through the Cancer Genome Atlas glioma collections~\cite{Bakas2017}, established the pre-operative glioma segmentation task and the ET/TC/WT composite-region evaluation convention still in use today. Successive editions have addressed increasingly challenging clinical scenarios, including pediatric tumors~\cite{Kazerooni2024}, and broader efforts~\cite{Maier2024} have sought to standardize the validation methodology to compare segmentation results across such heterogeneous benchmarks.

BraTS 2023~\cite{labella2023asnrmiccai, labella2024multi} introduced the task of intracranial meningioma segmentation, focusing on tumors that generally present well-defined boundaries but exhibit considerable variability in size, anatomical location, and imaging appearance. In contrast, BraTS 2024~\cite{BraTS24} addressed the considerably more challenging problem of post-treatment glioma segmentation. In this scenario, treatment-related changes such as resection cavities, edema, radiation effects, and necrotic tissue substantially increase anatomical variability and complicate the distinction between residual tumor and surrounding tissues. These complementary datasets provide an excellent opportunity to evaluate the robustness and generalization capability of modern segmentation architectures across clinically distinct conditions.

Although numerous studies have reported excellent performance for individual segmentation architectures, objective comparisons remain difficult because experimental protocols vary considerably across publications. Differences in preprocessing pipelines, data augmentation strategies, optimization methods, hyperparameter selection, training schedules, and hardware configurations frequently make it difficult to interpret the reported results~\cite{Dorfner2025, Missaoui2025}. Furthermore, many studies emphasize segmentation accuracy while providing only limited analysis of computational efficiency, despite its practical importance for clinical deployment.

To address these limitations, this work develops a unified experimental framework in which representative CNN-, Transformer-, and SSM-based architectures are trained and evaluated using identical evaluation procedures. Unlike previous studies that have focused on a single BraTS challenge, the proposed benchmark spans two clinically distinct datasets, allowing a comprehensive assessment of architectural robustness, segmentation accuracy, and computational efficiency.

\section{Materials and Methods}
\label{sec:materials_methods}

This section describes the datasets and experimental protocol used to construct a framework for automatic brain tumor segmentation. All models were trained using the same data splits, preprocessing steps, optimization settings, and inference procedures, ensuring that performance differences can be attributed to architectural design rather than to experimental factors.

\subsection{Brain Tumor Datasets}
\label{subsec:dataset}

\subsubsection{BraTS 2023 Intracranial Meningioma Challenge}

The first dataset~\cite{labella2023asnrmiccai} corresponds to the intracranial meningioma segmentation task introduced in the 2023 edition of the BraTS challenge. This edition extended the scope of BraTS beyond glioma segmentation, incorporating meningiomas, which are extra-axial tumors that often present well-defined boundaries but can vary considerably in size, shape, and anatomical location. Their tendency to appear in complex regions, such as the skull base, makes their automatic delineation a relevant and non-trivial segmentation problem.

\begin{figure}[ht!]
    \centering
    \includegraphics[width=1\textwidth]{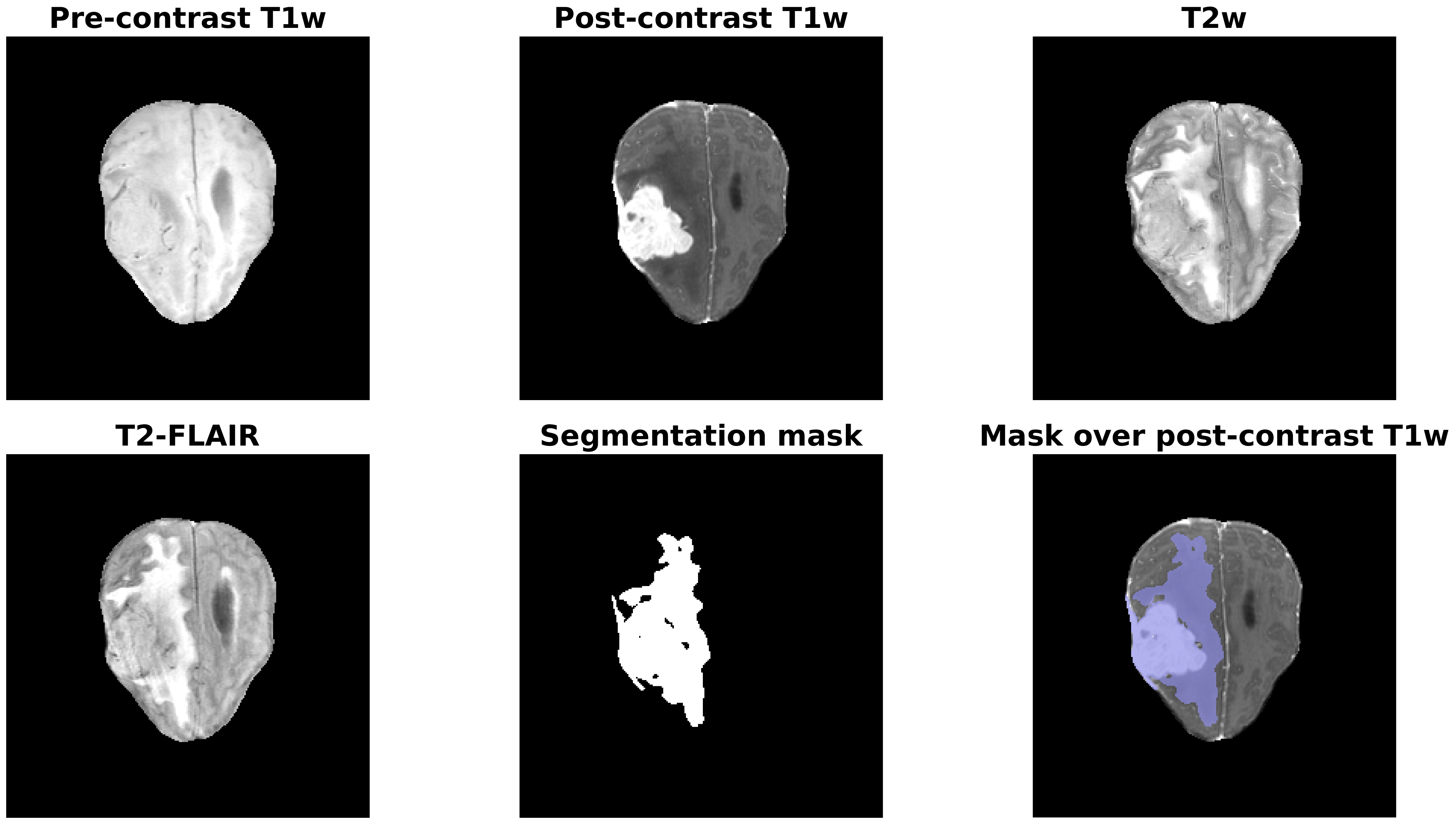}
    \caption{Patient case example (BraTS-MEN-00891-000) from the BraTS 2023 intracranial meningioma dataset showing the four MRI modalities (T1, T1c, T2, and T2-FLAIR), the corresponding tumor segmentation mask, and the segmentation mask overlaid on the post-contrast T1-weighted image.}
    \label{fig:brats_modalities_mask}
\end{figure}

Each patient study was acquired in a preoperative and pretreatment setting and included four co-registered MRI sequences: T1-weighted (T1), post-contrast T1-weighted (T1c), T2-weighted (T2), and T2-Fluid-Attenuated Inversion Recovery (T2-FLAIR). The corresponding segmentation mask is also provided for the annotated cases. Figure~\ref{fig:brats_modalities_mask} shows a representative example of the four modalities, the segmentation mask, and the mask overlaid on the post-contrast T1-weighted image. The BraTS preprocessing pipeline includes conversion from DICOM to the Neuroimaging Informatics Technology Initiative (NIfTI) format, rigid co-registration to a common anatomical space, resampling to an isotropic $1\times1\times1$~mm$^3$ resolution, and automatic skull-stripping. All volumes used in this work have spatial dimensions of $240 \times 240 \times 155$ voxels.

The dataset comprises 1\,141 cases, divided into 1\,000 training cases with expert-annotated segmentation masks and 141 official validation cases without reference masks. Since the validation partition does not include ground-truth annotations, we did not use it for the quantitative evaluation, and metrics were computed on the annotated training cases.

\begin{figure}[ht!]
    \centering
    \includegraphics[width=0.8\textwidth]{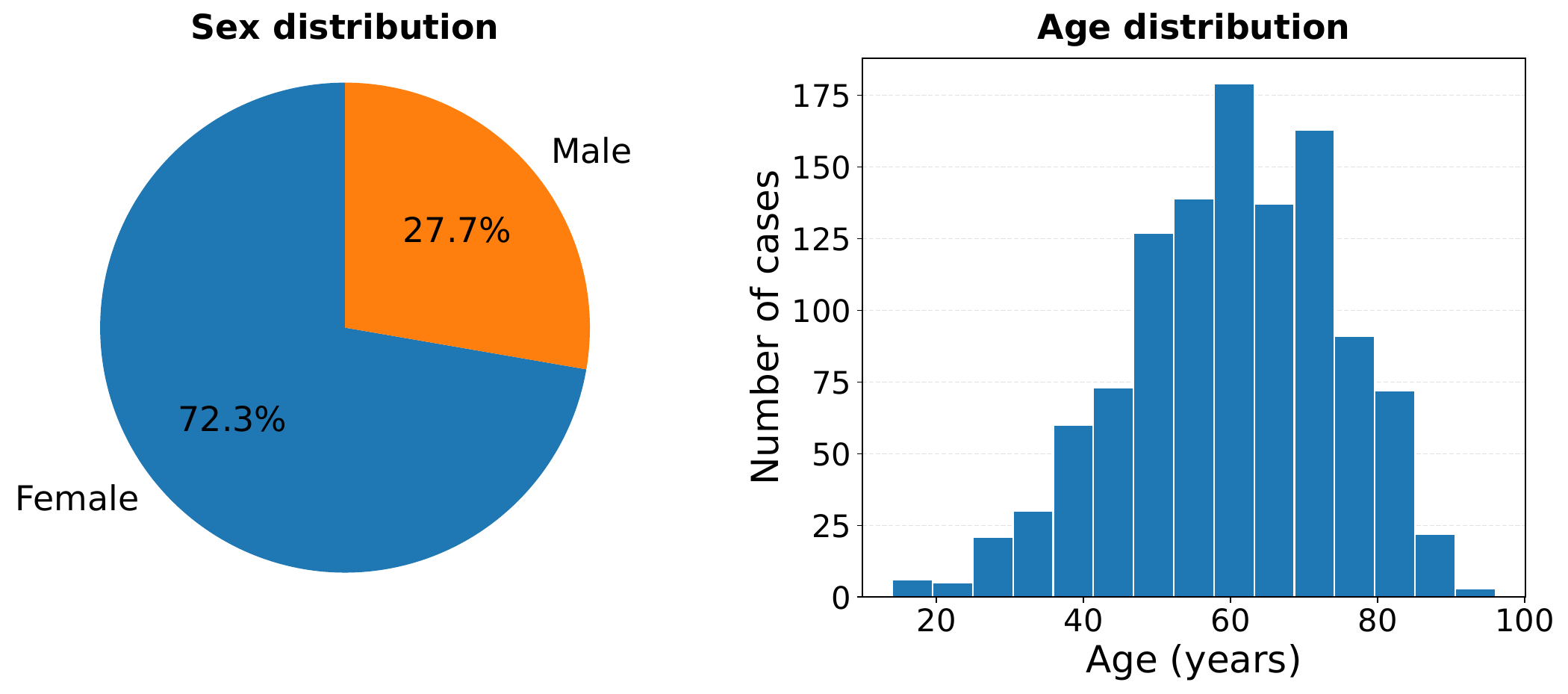}
    \caption{Distribution of patient sex and age in the BraTS 2023 dataset cohort.}
    \label{fig:brats_age_sex}
\end{figure}

\begin{figure}[ht!]
    \centering
    \includegraphics[width=0.6\textwidth]{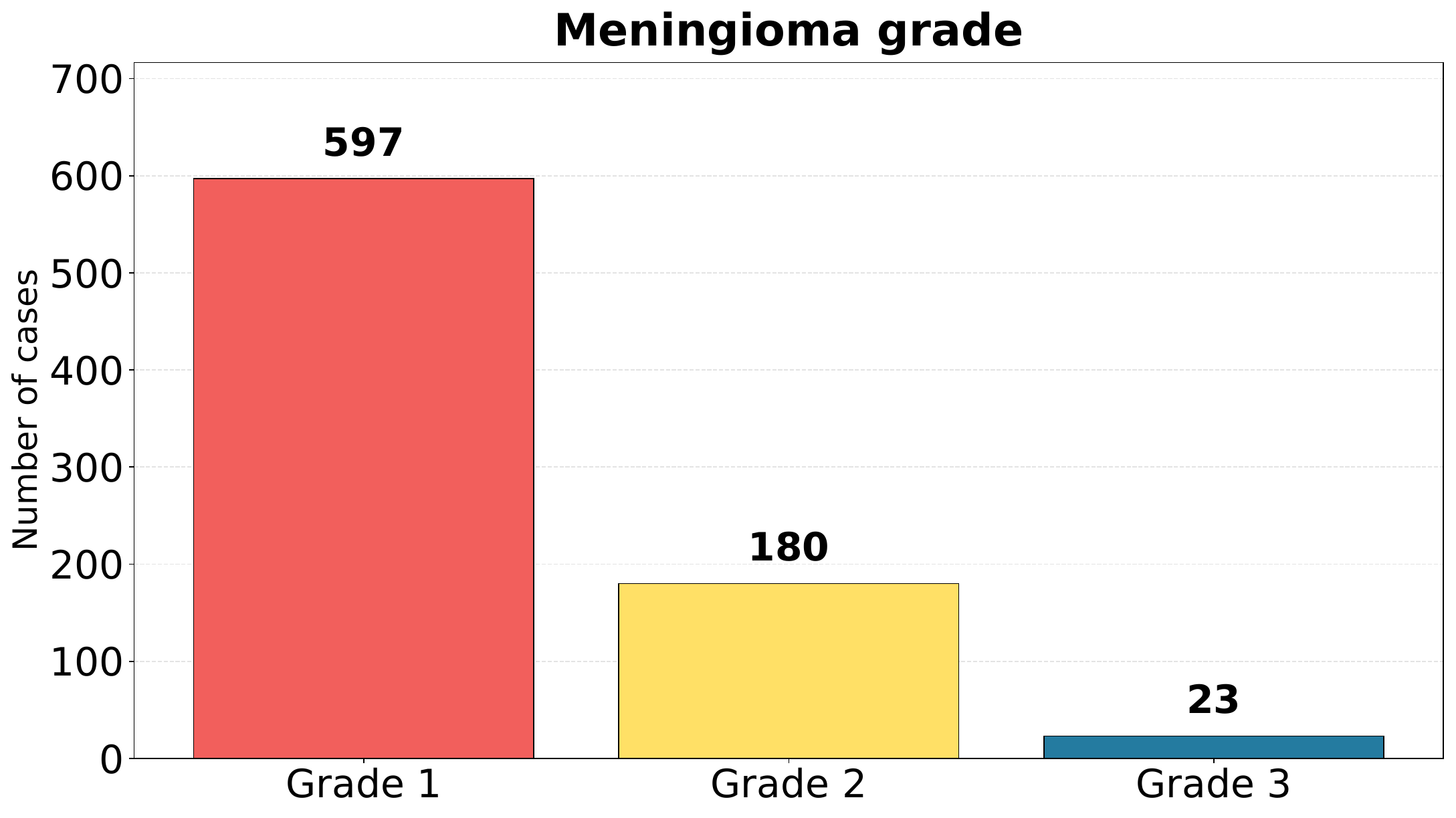}
    \caption{Distribution of meningioma grades in the BraTS 2023 dataset among the cases with available grade information.}
    \label{fig:brats_grade}
\end{figure}

The metadata for each patient includes patient age, sex, and meningioma grade at the time of imaging. Figure~\ref{fig:brats_age_sex} summarizes the sex and age distributions, showing a predominance of female patients (72.3\% of the cases), and a concentration of cases between 50 and 80 years of age. Figure~\ref{fig:brats_grade} further shows that, among the cases with available grade information---how aggressively the tumor behaves---, grade 1 meningiomas are the most frequent, whereas grades 2 and 3 are less represented. These clinical variables were not used during model training, but they help characterize the study population and indicate that the dataset is mainly composed of low-grade meningiomas.

\begin{figure}[ht!]
    \centering
    \includegraphics[width=0.9\textwidth]{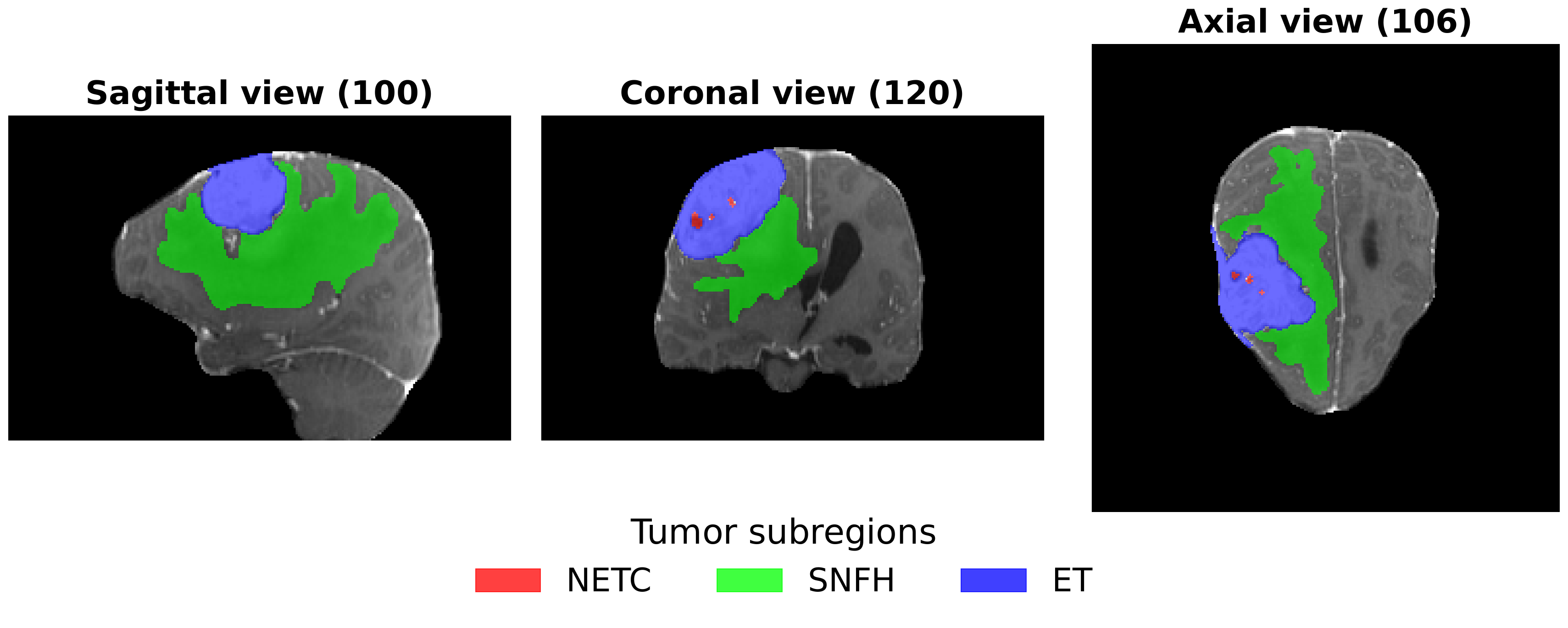}
    \caption{Patient case example (BraTS-MEN-00891-00) of the three tumor sub-regions annotated in BraTS 2023: enhancing tumor (ET), non-enhancing tumor core (NETC), and non-enhancing T2/FLAIR hyperintensity (SNFH).}
    \label{fig:brats_subregions}
\end{figure}

The segmentation masks distinguish three tumor sub-regions: enhancing tumor (ET), non-enhancing tumor core (NETC), and non-enhancing T2/FLAIR hyperintensity (SNFH). A representative visualization of these annotations is depicted in Figure~\ref{fig:brats_subregions}. Following the standard BraTS evaluation convention, the original labels were also combined into three composite regions: enhancing tumor (ET), tumor core (TC $=$ ET $\cup$ NETC), and whole tumor (WT $=$ ET $\cup$ NETC $\cup$ SNFH). 

\begin{figure}[ht!]
\subfloat[\centering]{\includegraphics[width=0.5\textwidth]{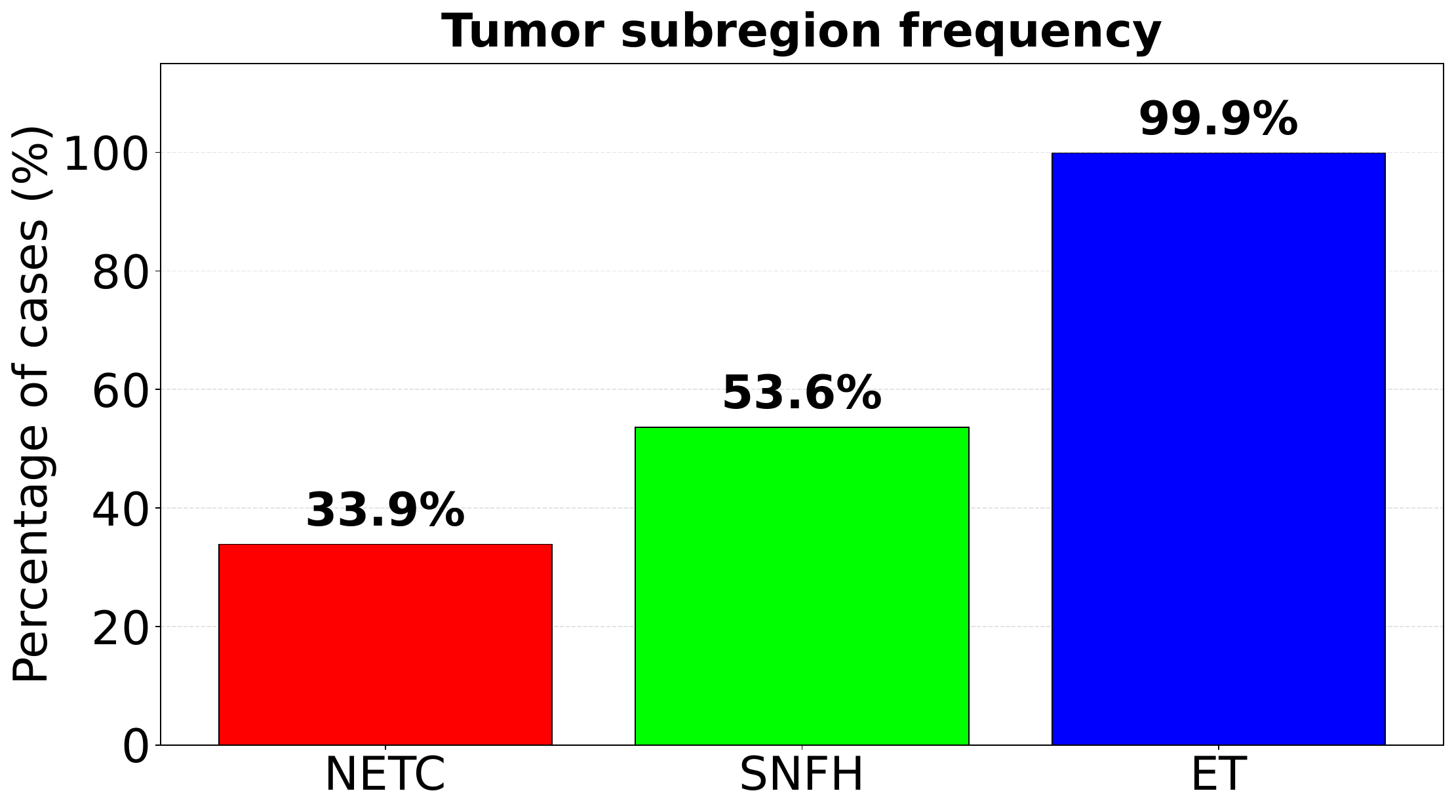}}
\subfloat[\centering]{\includegraphics[width=0.5\textwidth]{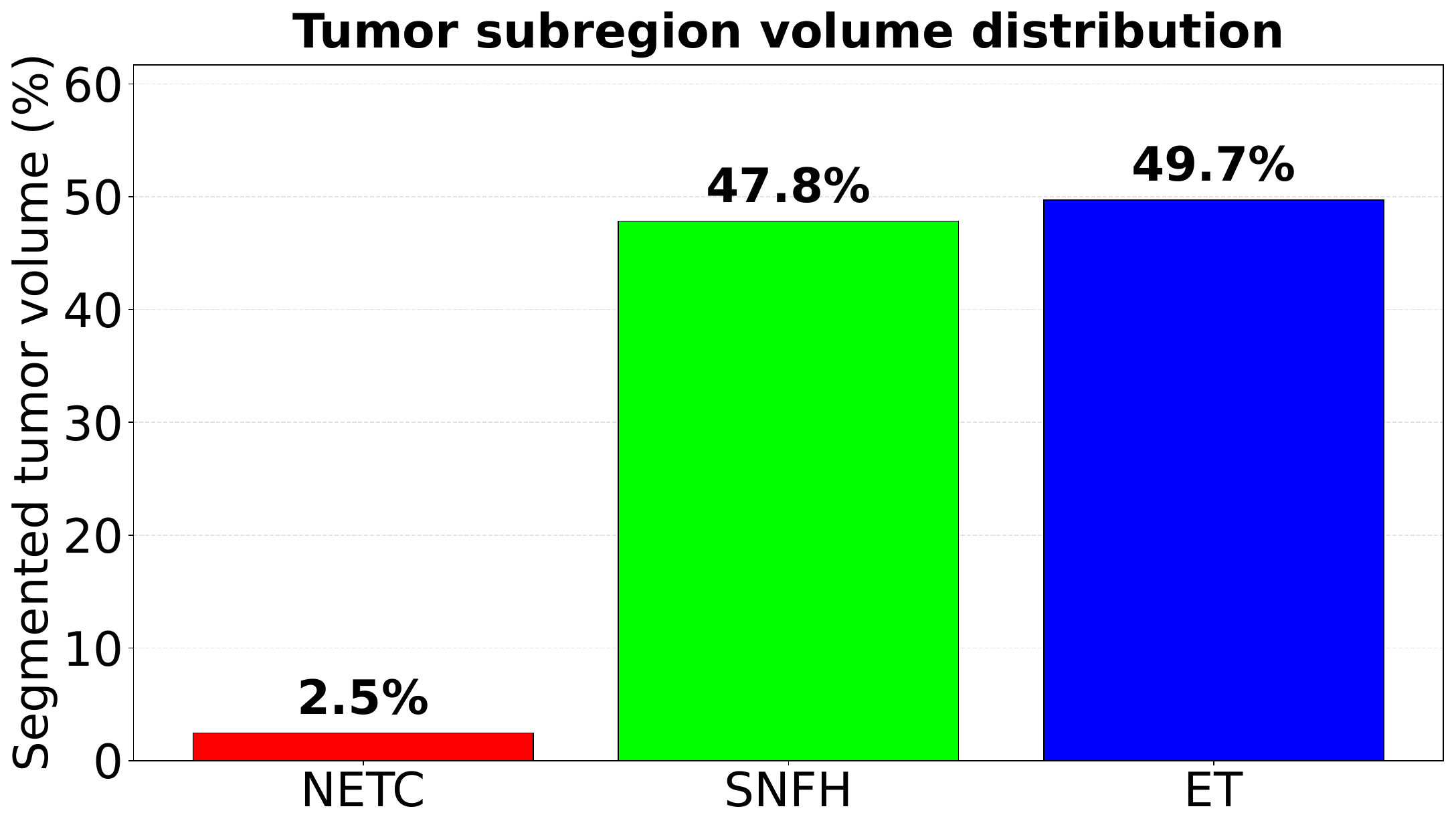}}
\caption{Analysis of the distribution of tumor sub-regions (NETC, SNFH, and ET) in the BraTS 2023 training cases: (\textbf{a}) Frequency of each annotated sub-region; (\textbf{b}) Percentage of each tumor sub-region regarding the total segmented tumor volume.}
\label{fig:brats_subregion_frequency}
\end{figure}

An exploratory analysis of the 1\,000 annotated cases revealed a significant imbalance among tumor sub-regions. In terms of occurrence, Figure~\ref{fig:brats_subregion_frequency} shows that ET appears in nearly all cases (99.9\%), while SNFH and NETC are present in 53.6\% and 33.9\% of cases, respectively. At the voxel level, the imbalance is also clear: ET and SNFH account for most of the segmented tumor volume, at 49.7\% and 47.8\%, respectively, whereas NETC represents only 2.5\%; see Figure~\ref{fig:brats_subregion_frequency} on the right. This heterogeneity was considered when selecting the loss function and interpreting the per-sub-region segmentation results.

\subsubsection{BraTS 2024 Post-Treatment Glioma Challenge}
\label{subsubsec:brats2024}

The second dataset~\cite{BraTS24} was obtained from the post-treatment glioma segmentation task of the 2024 BraTS challenge. Compared with preoperative segmentation settings, this task introduces a more complex radiological scenario, as surgical resection, therapy-induced tissue changes, edema, blood products, necrosis, and resection cavities can substantially alter normal anatomy, making the distinction between residual tumor and treatment-related changes particularly challenging.

\begin{figure}[ht!]
    \centering
    \includegraphics[width=1\textwidth]{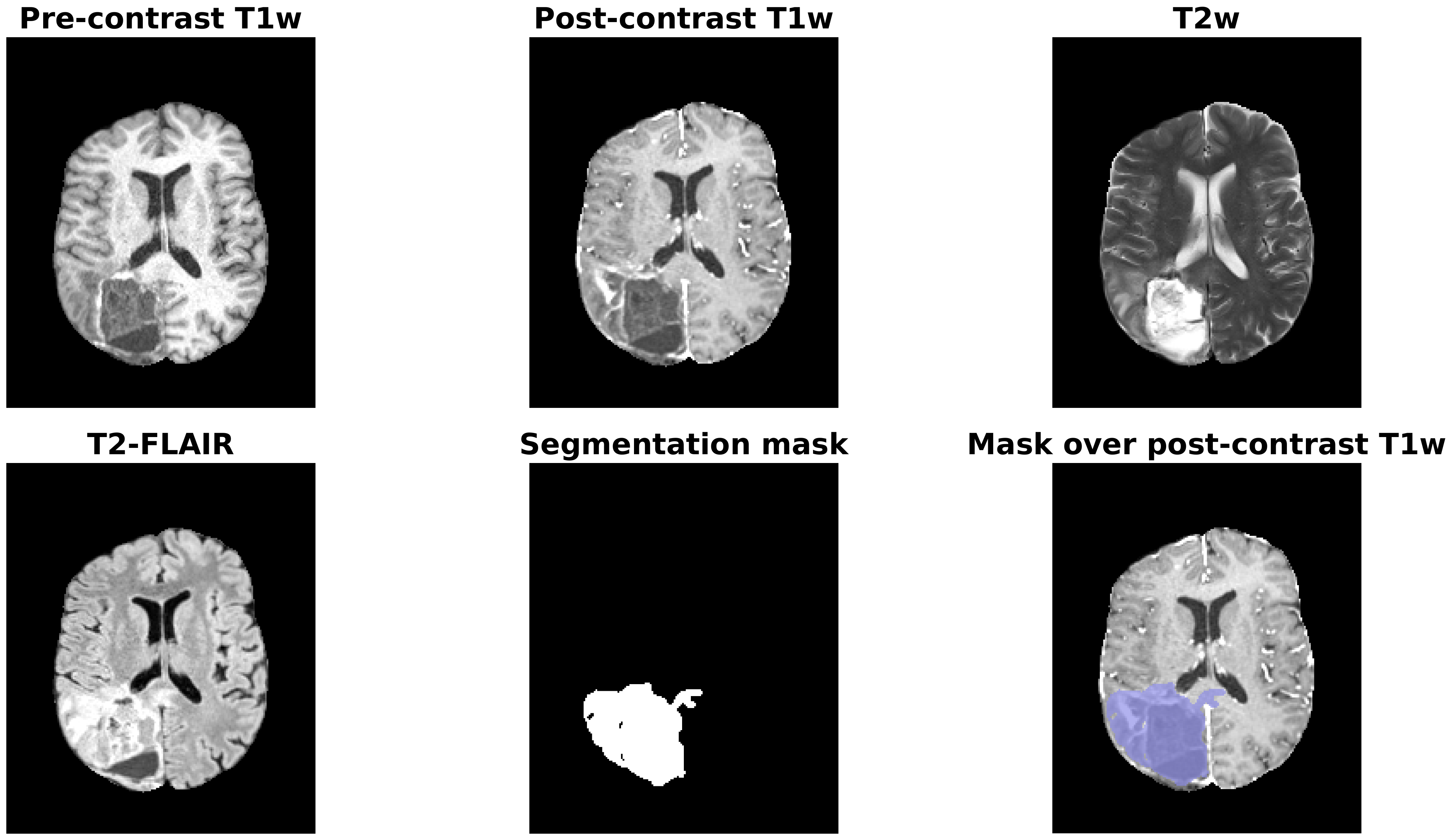}
    \caption{Patient case example (BraTS-GLI-03063-100) from the BraTS 2024 post-treatment dataset showing the four MRI modalities (T1, T1c, T2, and T2-FLAIR), the corresponding segmentation mask, and the segmentation mask overlaid on the post-contrast T1-weighted image.}
    \label{fig:brats2024_modalities_mask}
\end{figure}

The imaging data consist of four co-registered MRI sequences for each patient: T1-weighted (T1), post-contrast T1-weighted (T1c), T2-weighted (T2), and T2-FLAIR. These modalities provide complementary information about enhancement, edema, non-enhancing tumor components, and post-treatment anatomical alterations. Figure~\ref{fig:brats2024_modalities_mask} illustrates a representative case, including the four MRI modalities, the segmentation mask, and the overlay of the mask on the T1c image.

The dataset is divided into three main partitions: The two training folders (\textit{training\_data1\_v2} and \textit{training\_data\_additional}) contain 1\,350 and 271 patients, respectively, and both include MRI sequences together with their corresponding segmentation masks---the latter partition was released because corrections were introduced for some of the original cases---; the official validation set (\textit{validation\_data}) contains 188 patients, but only the MRI sequences are publicly available, so it was excluded from the quantitative evaluation due to the absence of reference masks. 

\begin{figure}[ht!]
    \centering
    \includegraphics[width=0.8\textwidth]{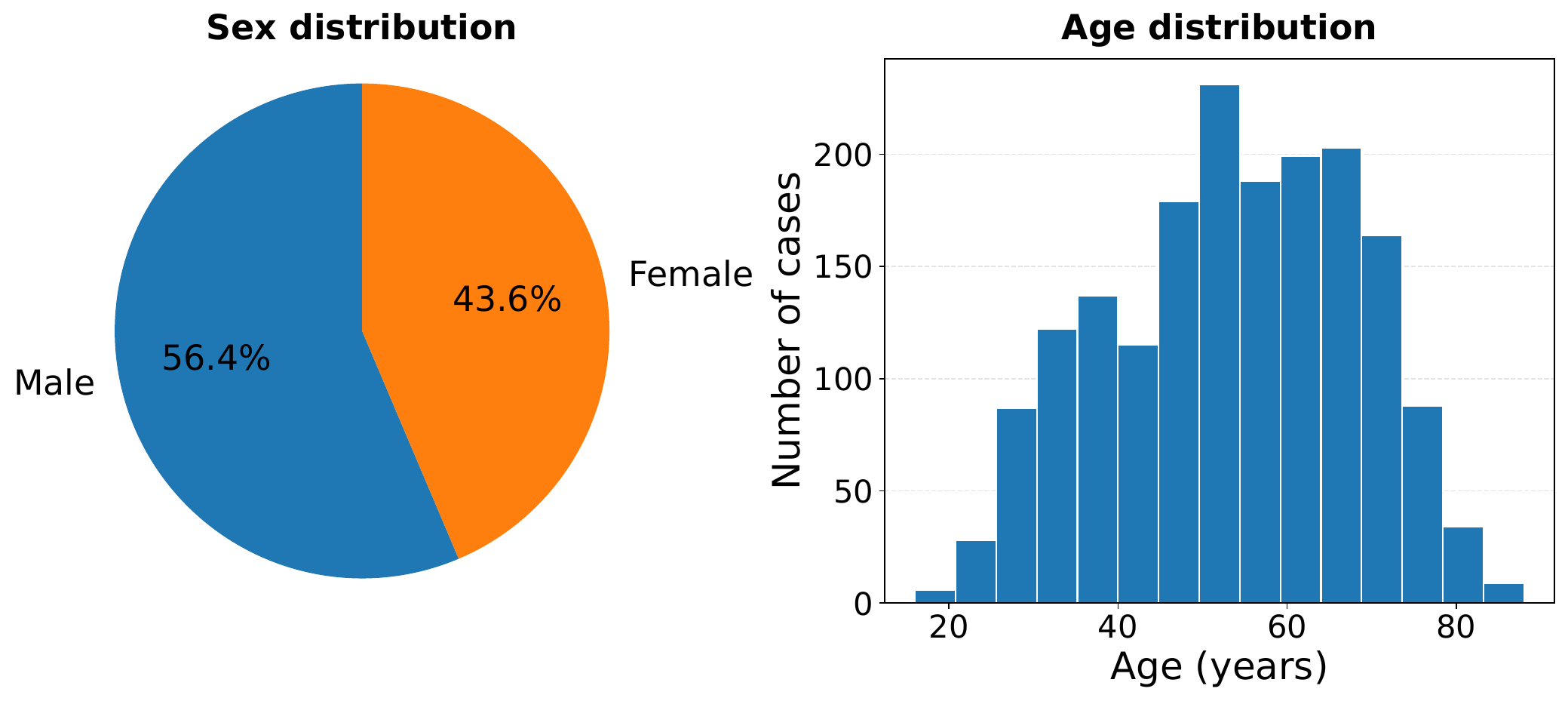}
    \caption{Distribution of patient sex and age in the BraTS 2024 dataset cohort.}
    \label{fig:brats2024_age_sex}
\end{figure}

\begin{figure}[ht!]
    \centering
    \includegraphics[width=0.6\textwidth]{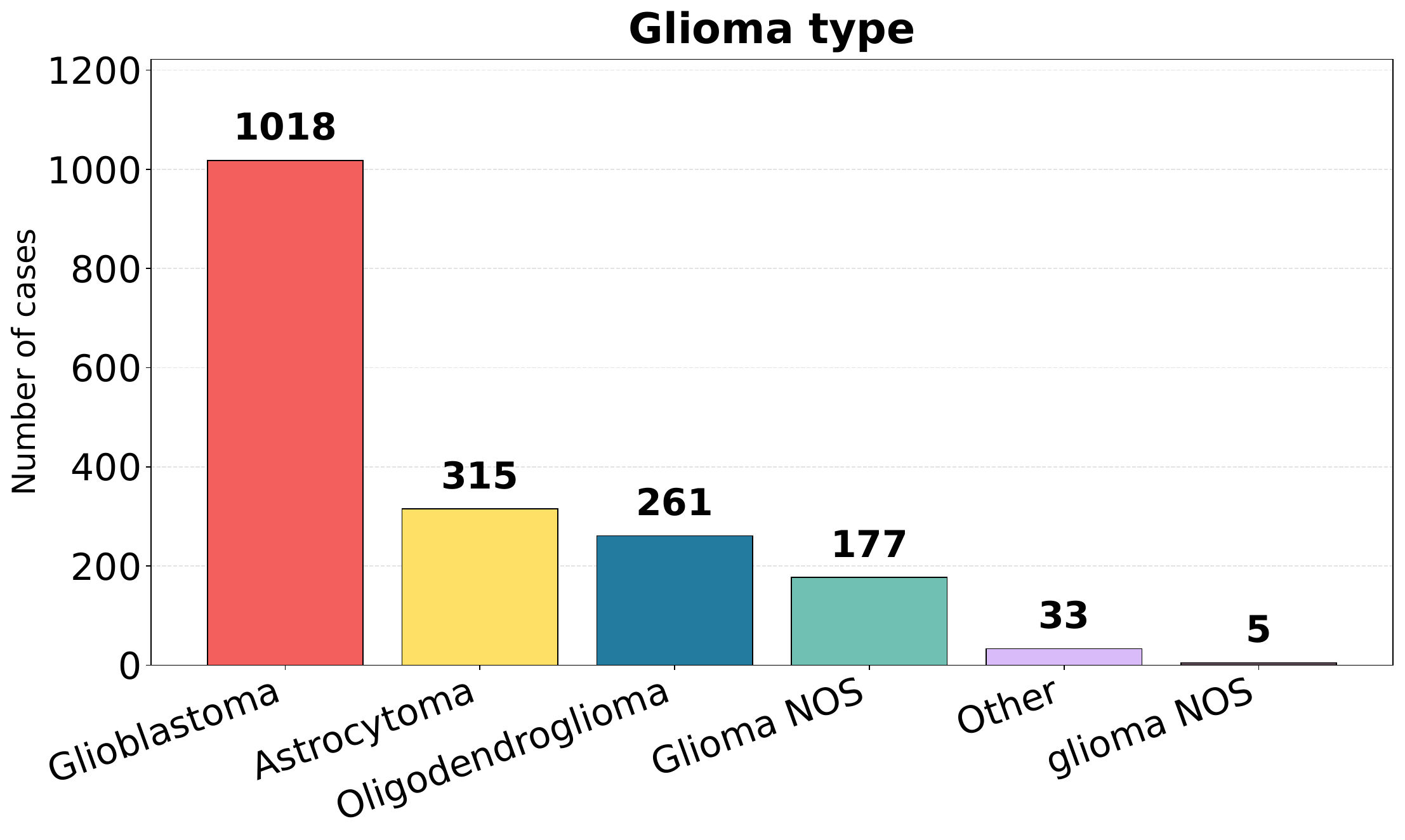}
    \caption{Distribution of glioma types in the BraTS 2024 dataset cohort.}
    \label{fig:brats2024_glioma_type}
\end{figure}

Patient-level metadata were also examined to characterize the cohort. As summarized in Figure~\ref{fig:brats2024_age_sex}, males represent a slightly larger proportion of the dataset (56.4\%) than females (43.6\%), and the age of most patients falls within approximately 45--70 years. Regarding diagnosis, Figure~\ref{fig:brats2024_glioma_type} shows that glioblastoma is the dominant tumor type, with 1\,018 cases, followed by astrocytoma, oligodendroglioma, and glioma NOS. The remaining categories are considerably less frequent. 

\begin{figure}[ht!]
    \centering
    \includegraphics[width=0.9\textwidth]{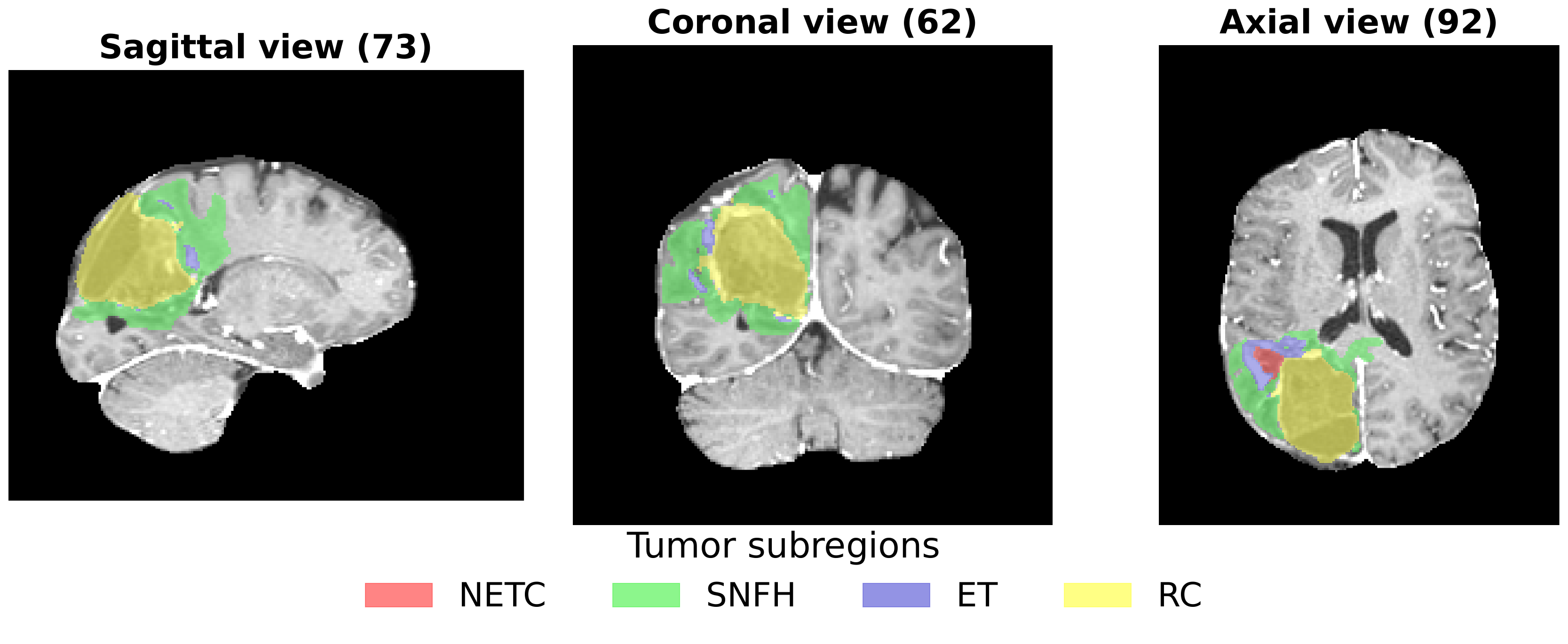}
    \caption{Patient case example (BraTS-GLI-03063-100) of post-contrast T1w of the four tumor sub-regions annotated in BraTS 2024: non-enhancing tumor core (NETC), surrounding non-enhancing FLAIR hyperintensity (SNFH), enhancing tumor (ET), and resection cavity (RC).}
    \label{fig:brats2024_subregions}
\end{figure}

Unlike the BraTS 2023 meningioma task, the BraTS 2024 annotations include four post-treatment glioma sub-regions: non-enhancing tumor core (NETC), surrounding non-enhancing FLAIR hyperintensity (SNFH), enhancing tumor (ET), and resection cavity (RC). These labels reflect the coexistence of tumor-related tissue and treatment-induced changes within the same volume. Figure~\ref{fig:brats2024_subregions} shows a representative multiplanar visualization of the annotated sub-regions.

\begin{figure}[ht!]
\subfloat[\centering]{\includegraphics[width=0.5\textwidth]{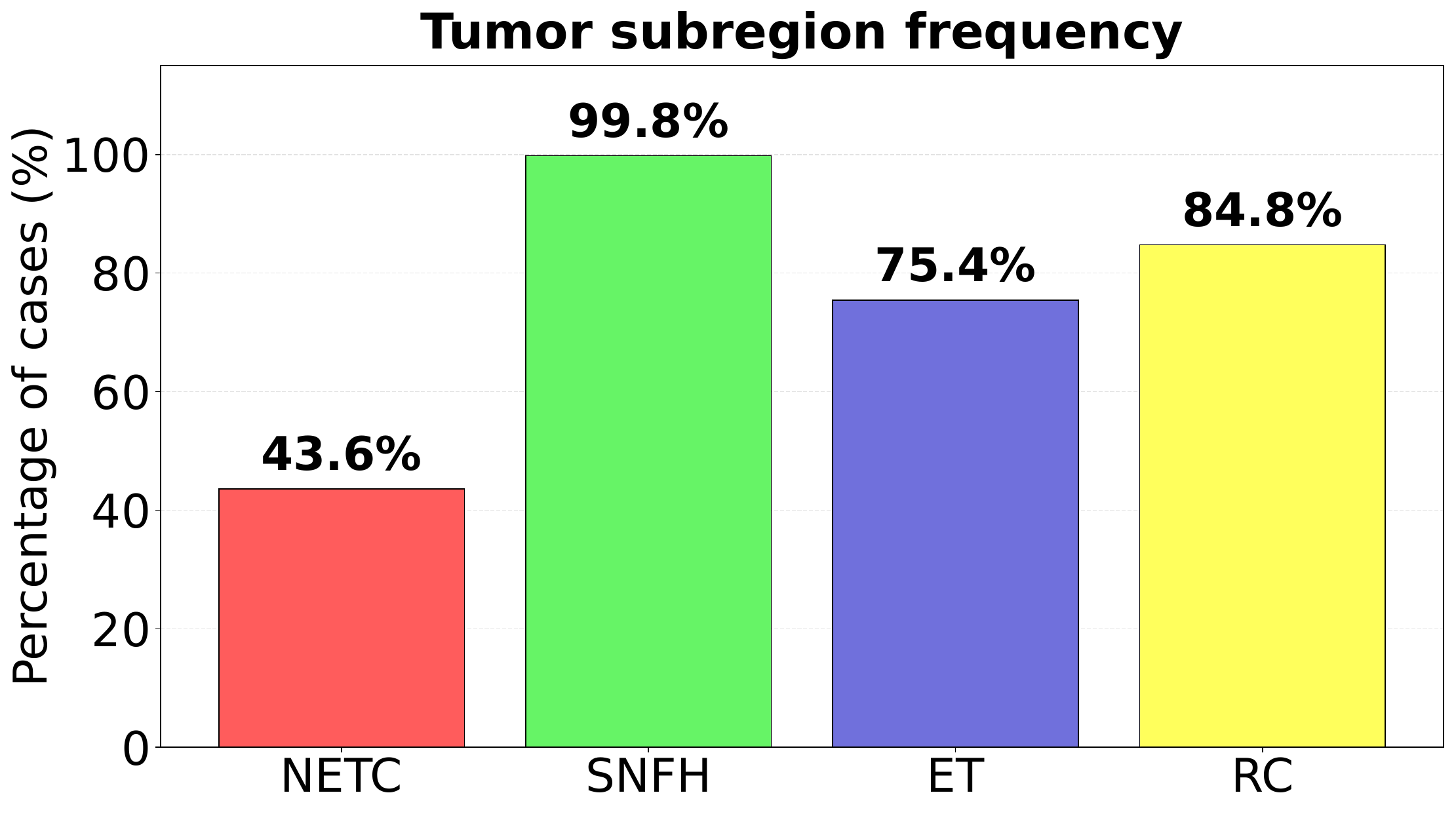}}
\subfloat[\centering]{\includegraphics[width=0.5\textwidth]{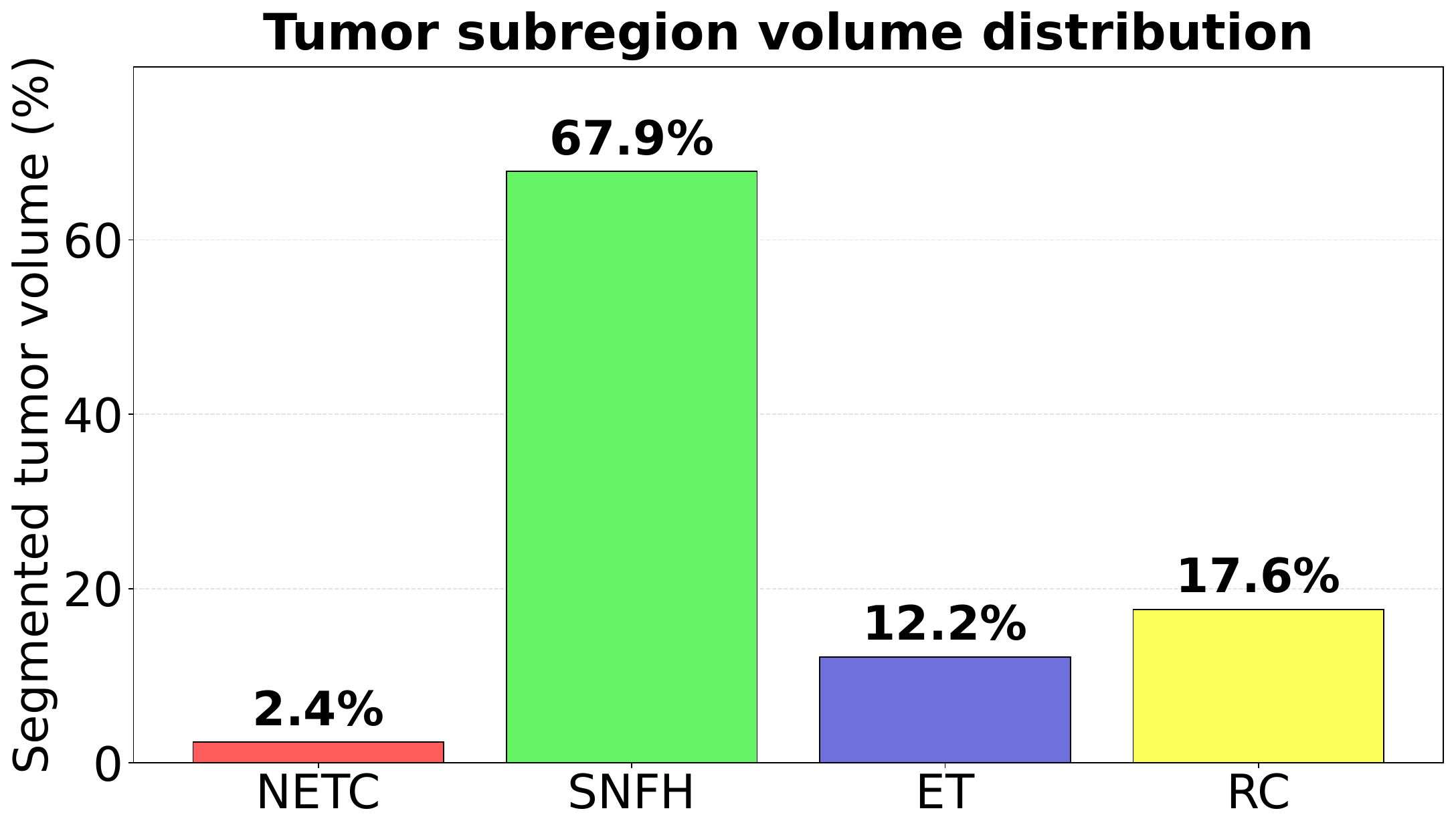}}
\caption{Analysis of the distribution of tumor sub-regions (NETC, SNFH, ET, and RC) in the BraTS 2024 training cases: (\textbf{a}) Frequency of each annotated sub-region; (\textbf{b}) Percentage of each tumor sub-region regarding the total segmented tumor volume.}
\label{fig:brats2024_subregion_frequency}
\end{figure}

The annotated training cases also present a strong imbalance across sub-regions. Figure~\ref{fig:brats2024_subregion_frequency} shows that SNFH is present in almost all patients (99.8\%), followed by RC (84.8\%), ET (75.4\%), and NETC (43.6\%). The volumetric contribution of each sub-region is similarly uneven: SNFH accounts for 67.9\% of the segmented tumor volume, RC for 17.6\%, ET for 12.2\%, and NETC for only 2.4\%; see Figure~\ref{fig:brats2024_subregion_frequency} on the right. Therefore, NETC is both the least frequent sub-region and the smallest in terms of annotated volume, making it particularly vulnerable to class imbalance and segmentation errors.

All BraTS 2024 volumes analyzed in this work have spatial dimensions of $182 \times 218 \times 182$ voxels and an isotropic voxel spacing of $1.0 \times 1.0 \times 1.0$~mm$^3$. In addition to the imbalance among tumor sub-regions, approximately 98.99\% of voxels correspond to background, emphasizing the strong foreground-background imbalance inherent to volumetric brain tumor segmentation. These properties were taken into account when defining the preprocessing strategy, training-time cropping, loss function, and interpretation of the per-sub-region results.

\subsubsection{Data Partitioning and Preprocessing}
\label{subsubsec:partitioning}

The official validation partitions of the datasets do not provide segmentation masks; thus, the quantitative evaluation was performed using only the annotated training data. For BraTS 2023, the experimental splits were derived from the official training folder, and for BraTS 2024, the annotated cases from \textit{training\_data1\_v2} and \textit{training\_data\_additional} were considered jointly. 

A five-fold cross-validation strategy was adopted to obtain a more robust comparison between architectures. In each iteration, one fold was reserved for testing, corresponding to approximately 20\% of the annotated cases. The remaining four folds were used for model training: 90\% of these cases were assigned to training, and the remaining 10\% to internal validation. Therefore, each model was trained, validated, and tested multiple times using different test partitions.

The same data partitions were reused across all architectures, ensuring that every model was evaluated under identical data conditions. This design reduces the dependence of the results on a single train/test split and allows for a fair comparison between the evaluated models.

A common preprocessing and augmentation pipeline was implemented using the Medical Open Network for Artificial Intelligence (MONAI)~\cite{cardoso2022monai} library. The same pipeline was applied to both BraTS 2023 and BraTS 2024 to ensure homogeneous input preparation across datasets and architectures. The original spatial dimensions depend on the dataset, but all cases were standardized through the same sequence of transformations. The pipeline comprised the following steps:

\begin{enumerate}
    \item Data loading and channel arrangement. The four MRI modalities and the corresponding segmentation mask were loaded from NIfTI files. The image channels were arranged in channel-first format, and data were converted to tensor-compatible types.
    \item Geometric standardization. All volumes were reoriented to the RAS anatomical convention and resampled to an isotropic voxel spacing of $1.0 \times 1.0 \times 1.0$\,mm$^3$. The foreground region was then cropped, and spatial padding was applied to ensure a minimum spatial size of $128 \times 128 \times 128$\,voxels.
    \item Intensity normalization. Voxel intensities were normalized independently for each MRI modality, so that all input channels contributed to a comparable numerical scale.
    \item Data augmentation. For the training subset only, random spatial crops were extracted, followed by random flipping along the three spatial axes, random intensity scaling, and random intensity shifting. Spatial transformations were applied consistently to the image channels and to the segmentation mask to preserve voxel-wise correspondence.
    \item Validation/test-time preparation. For validation and testing, no random augmentation was applied. Only the deterministic preprocessing steps were used before the sliding-window inference strategy.
\end{enumerate}

\subsection{Neural Network Architectures}
\label{subsec:architectures}

Five three-dimensional segmentation architectures were evaluated under the same experimental framework: Three consolidated reference models (3D U-Net, SegResNet, and Swin~UNETR) and two more recent proposals from the Mamba family of SSMs (SegMamba and SegMambaV2). All architectures receive a four-channel input (one channel per MRI modality) and produce voxel-wise outputs with four classes for BraTS 2023 (background plus the three tumor sub-regions) and five classes for BraTS 2024 (background plus the four tumor sub-regions). Table~\ref{tab:architectures} summarizes the underlying design principles of each model.

\begin{table}[ht!]
\centering
\caption{Overview of the segmentation architectures evaluated in this study.}
\label{tab:architectures}
\begin{tabular}{llp{7cm}}
\toprule
\textbf{Model} & \textbf{Family} & \textbf{Core idea} \\
\midrule
U-Net~3D~\cite{cciccek20163d} & Convolutional & Volumetric extension of U-Net with a symmetric encoder--decoder and skip connections. \\
SegResNet~\cite{SegResNet} & Residual Convolutional & Asymmetric encoder--decoder with residual blocks and an auxiliary variational auto-encoder branch for regularization. \\
Swin~UNETR~\cite{Swin_UNETR} & Hybrid Transformer & Hierarchical Swin Transformer encoder with shifted-window self-attention coupled to a convolutional decoder. \\
SegMamba~\cite{xing2024segmamba} & State-Space Model & U-shaped network with tri-oriented Mamba blocks for near-linear-complexity long-range dependency modeling. \\
SegMambaV2~\cite{xing2025segmamba} & State-Space Model & Hybrid convolution/Mamba architecture with hierarchical-scale downsampling and ortho-oriented Mamba blocks. \\
\bottomrule
\end{tabular}
\end{table}

\subsubsection{3D U-Net}

3D U-Net~\cite{cciccek20163d} extends the original two-dimensional U-Net to volumetric data by replacing every operation with its three-dimensional counterpart. The network follows an encoder-decoder topology: the encoder path progressively reduces the spatial resolution while increasing the number of feature representations through successive $3\times3\times3$ convolutions, batch normalization, and ReLU activations,

\begin{equation}
    \hat{y} = \mathrm{ReLU}\left(\mathrm{BatchNorm}\left(\mathrm{Conv3D}(x)\right)\right),
\end{equation}
followed by spatial down-sampling, while the decoder path applies transposed convolutions to progressively recover the original resolution. Skip connections concatenate encoder feature maps with decoder feature maps of matching resolution, allowing fine-grained spatial information to be recovered during reconstruction. In this work, the network was configured with a five-level encoder-decoder hierarchy. The encoder progressively reduced the spatial resolution while increasing the number of learned feature representations, whereas the decoder gradually reconstructed the original resolution through transposed convolutions.

\subsubsection{SegResNet}

SegResNet~\cite{SegResNet} adopts an asymmetric encoder–decoder architecture in which the encoder, composed of residual blocks with group normalization, is substantially deeper than the decoder. Each residual block can be summarized as
\begin{align}
    h_1 &= \mathrm{ReLU}(\mathrm{GN}(x)), \nonumber    \\  
    h_2 &= \mathrm{Conv3D}(h_1), \nonumber    \\
    h_3 &= \mathrm{ReLU}(\mathrm{GN}(h_2)), \nonumber    \\  
    h_4 &= \mathrm{Conv3D}(h_3), \nonumber    \\
    \hat{y} &= x + h_4, 
\end{align}
where the residual connection facilitates gradient propagation and enables the training of deeper networks. Group normalization (GN) is employed instead of batch normalization because volumetric segmentation often relies on very small batch sizes due to GPU memory limitations. The original architecture also incorporates an auxiliary variational auto-encoder (VAE) branch connected to the deepest encoder stage, which reconstructs the input volume during training to provide additional regularization. In this work, the VAE branch was omitted to reduce memory consumption while preserving the main segmentation architecture.

\subsubsection{Swin UNETR}

Swin~UNETR~\cite{Swin_UNETR} combines a hierarchical Swin Transformer~\cite{liu2021swin,liu2022swin} encoder with a convolutional decoder connected through skip connections. The input volume is first partitioned into non-overlapping $2\times2\times2$ patches, which are linearly embedded into feature tokens. Self-attention is computed locally within non-overlapping three-dimensional windows (Window Multi-head Self-Attention, W-MSA),
\begin{equation}
    \mathrm{Attention}(Q,K,V) = \mathrm{Softmax}\!\left(\frac{QK^{T}}{\sqrt{d}}\right)V,
\end{equation}
and windows are subsequently shifted by half their size (Shifted-Window Multi-head Self-Attention, SW-MSA) to enable cross-window interactions, following the two-step block
\begin{align}
    \hat{z}^{l} &= \mathrm{W\text{-}MSA}(\mathrm{LN}(z^{l-1})) + z^{l-1}, &
    z^{l} &= \mathrm{MLP}(\mathrm{LN}(\hat{z}^{l})) + \hat{z}^{l}, \nonumber \\
    \hat{z}^{l+1} &= \mathrm{SW\text{-}MSA}(\mathrm{LN}(z^{l})) + z^{l}, &
    z^{l+1} &= \mathrm{MLP}(\mathrm{LN}(\hat{z}^{l+1})) + \hat{z}^{l+1}.
\end{align}

A patch-merging operation progressively reduces the spatial resolution while increasing the feature dimensionality, producing a hierarchical feature pyramid. The convolutional decoder reconstructs the segmentation mask by progressively up-sampling and fusing encoder features through skip connections, using residual convolutional blocks with instance normalization.

\subsubsection{SegMamba}

SegMamba~\cite{xing2024segmamba} replaces the global self-attention mechanism of Transformer-based models with structured SSMs from the Mamba family, which capture long-range dependencies with near-linear computational complexity. The encoder is built from Tri-oriented Spatial Mamba (TSMamba) blocks, each combining a Gated Spatial Convolution (GSC) module, which preserves local three-dimensional context before the sequence is flattened, with a Tri-oriented Mamba (ToM) module, which models global dependencies along forward, reverse, and inter-slice directions and sums the resulting representations. Skip connections incorporate a Feature-level Uncertainty Estimation (FUE) module that attenuates low-confidence encoder features, computed as
\begin{equation}
    \bar{z}^{i} = \sigma\!\left(\frac{1}{C_i}\sum_{c=1}^{C_i} z_{c}^{i}\right), \qquad
    u^{i} = -\bar{z}^{i}\log(\bar{z}^{i}), \qquad
    \tilde{z}^{i} = z^{i} + z^{i}\cdot(1-u^{i}),
\end{equation}
before fusion with the up-sampled decoder features.

The implementation used in this work was adapted from the authors' repository. The publicly available code differs slightly from the architecture described in the original publication: the GSC module combines feature branches through addition rather than the gated multiplication described in the paper; the FUE module is not included in the released implementation, and the decoder relies on standard convolutional building blocks. Apart from these implementation differences, the network follows the original four-stage hierarchical design, progressively increasing the feature representations as the spatial resolution decreases.

\subsubsection{SegMambaV2}

SegMambaV2~\cite{xing2025segmamba} is a four-stage hybrid evolution of SegMamba designed to mitigate the loss of fine spatial detail during down-sampling and to enrich the modeling of global context. The first two stages employ depth-wise convolutional blocks for efficiently extracting high-resolution local anatomical features, while the last two stages employ enhanced TSMamba blocks. The standard stem layer is replaced by a Hierarchical Scale Downsampling (HS-Downsampling) module, which combines three parallel branches with different receptive fields, fused through a convolutional layer to preserve multi-scale spatial detail during resolution reduction. The Tri-oriented Mamba module is redesigned as a Tri-orientated Ortho Mamba (ToOM) module that models dependencies separately along the three anatomical planes (coronal, sagittal, axial) and combines them by direct summation,

\begin{equation}
    \mathrm{ToOM}(z) = \mathrm{ToM\text{-}Coronal}(z) + \mathrm{ToM\text{-}Sagittal}(z) + \mathrm{ToM\text{-}Axial}(z).
\end{equation}

As with SegMamba, the implementation was adapted from the authors' repository; no dependency-related modifications were required. Consistent with the architecture description, the released code approximates HS-Downsampling with a single three-dimensional convolutional layer, implements ToOM via axis permutation combined with one-dimensional Mamba layers, replaces the GSC module with large-kernel convolutions and a channel-attention mechanism, and delegates decoding to standard convolutional building blocks. The network follows the original four-stage hierarchical design and employs stochastic depth regularization during training. Because the preliminary training of this model exhibited unstable, exploding gradients, gradient clipping was additionally applied during optimization.

\subsection{Training Configuration and Optimization}
\label{subsec:optimization}

All architectures were trained from scratch rather than fine-tuned from publicly available pretrained checkpoints. This decision was adopted to ensure a fair comparison under the same experimental conditions and to avoid potential data leakage. Since the official validation partitions do not provide reference masks, the quantitative evaluation was therefore performed using splits derived from the annotated training data.

Model parameters were optimized using the Dice Cross-Entropy Loss function, which combines the multi-class Dice loss~\cite{milletari2016v} with the cross-entropy loss. This choice was motivated by the pronounced foreground-background and inter-class imbalance observed in the datasets described in Section~\ref{subsec:dataset}. The Dice component promotes spatial overlap between the predicted and reference masks, whereas the cross-entropy component penalizes voxel-wise classification errors.

For a volume with $N$ voxels and $C$ classes, let $\hat{y}_{i,c}$ denote the predicted softmax probability for voxel $i$ belonging to class $c$, and let $y_{i,c}$ denote the corresponding one-hot encoded ground-truth label. Incorporating a smoothing factor $\epsilon$ to improve numerical stability, the Dice loss is defined as

\begin{equation}
    \mathcal{L}_{\mathrm{Dice}} = 1 -
        \frac{1}{|C|}\sum_{c \in C} \frac{2\sum_{i=1}^{N} \hat{y}_{i,c} y_{i,c} + \epsilon}
                                         {\sum_{i=1}^{N} \hat{y}_{i,c}+\sum_{i=1}^{N} y_{i,c}+ \epsilon}.
    \label{eq:diceloss}
\end{equation}

The cross-entropy component is defined as

\begin{equation}
    \mathcal{L}_{\mathrm{CE}} = -\frac{1}{N} \sum_{i=1}^{N} \sum_{c \in C} y_{i,c}\log(\hat{y}_{i,c}),
    \label{eq:celoss}
\end{equation}
where the summation is performed over all voxels and classes. The final training objective combines both expressions:

\begin{equation}
    \mathcal{L}_{\mathrm{DiceCE}} = \mathcal{L}_{\mathrm{Dice}} + \mathcal{L}_{\mathrm{CE}}.
    \label{eq:dicece}
\end{equation}

The background class was included in the loss computation across all experiments, consistent with common practice for multi-class segmentation losses such as DiceCELoss.

Parameters were updated using the AdamW optimizer~\cite{AdamW}, with a weight decay of $1\times10^{-5}$. A learning rate of $1\times10^{-4}$ was used for most architectures. For SegMamba and SegMambaV2, the learning rate was reduced to $1\times10^{-5}$, and gradient clipping with a maximum norm of 1 was applied to improve training stability and mitigate gradient explosion observed during preliminary experiments. Automatic mixed-precision training was used throughout to reduce GPU memory consumption and accelerate computation.

Table~\ref{tab:hyperparameters} summarizes the main training hyperparameters. Whenever possible, these settings were kept identical across architectures to ensure that performance differences were mainly attributable to the model design rather than to the training configuration. The batch size was set to 1 due to the memory requirements of three-dimensional MRI volumes.

\begin{table}[ht!]
\centering
\caption{Training hyperparameters used in the experiments.}
\label{tab:hyperparameters}
\begin{tabular}{lc}
\toprule
\textbf{Hyperparameter} & \textbf{Value} \\
\midrule
Training crop size & $128\times128\times128$ voxels \\
Batch size & 1 \\
Sliding-window batch size & 2 \\
Maximum epochs & 100 \\
Validation interval & every 5 epochs \\
Optimizer & AdamW \\
Weight decay & $1\times10^{-5}$ \\
General learning rate & $1\times10^{-4}$ \\
SegMamba and SegMambaV2 learning rate & $1\times10^{-5}$ \\
SegMamba and SegMambaV2 gradient clipping & maximum norm 1 \\
Loss function & Dice + cross-entropy loss \\
Precision & Automatic mixed precision \\
Number of workers & 4 \\
SegMamba and SegMambaV2 number of workers & 2 \\
\bottomrule
\end{tabular}
\end{table}

Model performance was monitored every 5 epochs using the held-out validation subset of each cross-validation fold, without stochastic data augmentation. Validation inference was performed using the sliding-window strategy, with a window size of $128\times128\times128$ voxels and a sliding-window batch size of 2. Predicted probabilities were converted to discrete labels, and validation metrics were computed on the resulting segmentation masks.

The mean Dice coefficient across the evaluated BraTS regions was used as the model-selection criterion. Whenever a new best validation Dice score was obtained, the model state, optimizer state, epoch index, and best validation metric were saved. After training, this best-performing checkpoint was restored and used for the final evaluation on the corresponding test split.

\subsection{Evaluation Protocol}
\label{subsec:evaluation}

Test-time inference reused the best checkpoint obtained during validation and followed the same sliding-window strategy applied during validation. Each padded volume was processed using sliding-window inference with overlapping $128 \times 128 \times 128$-voxel windows, a 25\% overlap ratio, and the sliding-window batch size defined in the experimental configuration. Window-level predictions were aggregated to reconstruct the complete volumetric model output.

The aggregated predictions were converted into discrete segmentation maps by selecting the maximum probability channel for each voxel. The resulting label maps were stored as NIfTI files together with the corresponding reference segmentations and subsequently used for the final evaluation.

The evaluation protocol provides a comprehensive assessment of segmentation performance using standard overlap and boundary-distance measurements. Specifically, the Dice similarity coefficient, Intersection over Union (IoU), and 95th-percentile Hausdorff Distance (HD95) were computed using MONAI metrics. All metrics were calculated on a per-region basis, excluding the background class, and subsequently averaged to obtain global scores.

The evaluated regions followed the definitions established by each BraTS challenge. For the BraTS 2023 dataset, the individual tumor sub-regions were NETC, SNFH, and ET, together with the composite regions ET, TC, and WT. For the BraTS 2024 dataset, the evaluated individual classes were NETC, SNFH, ET, and RC. The composite regions ET, TC, and WT were additionally evaluated according to the challenge definitions, with RC excluded from these composite regions.

The Dice similarity coefficient measures volumetric overlap between the predicted segmentation $\hat{y}$ and the reference segmentation $y$:
\begin{equation}
    \mathrm{Dice}(\hat{y},y) = \frac{2\,|\hat{y} \cap y|}{|\hat{y}| + |y|}.
\end{equation}

Intersection over Union (IoU) provides a stricter overlap measure by normalizing the intersection with respect to the union of both regions:
\begin{equation}
    \mathrm{IoU}(\hat{y},y) = \frac{|\hat{y} \cap y|}{|\hat{y} \cup y|}.
\end{equation}

The 95th-percentile Hausdorff Distance (HD95) evaluates the agreement between segmentation boundaries. It is derived from the classical Hausdorff distance:
s\begin{equation}
    \mathrm{HD}(\hat{y},y) =
    \max\left\{
    \sup\nolimits_{a \in \hat{y}} \inf\nolimits_{b \in y} d(a,b),
    \sup\nolimits_{b \in y} \inf\nolimits_{a \in \hat{y}} d(b,a)
    \right\},
\end{equation}
where the 95th percentile is used instead of the maximum distance to reduce sensitivity to isolated outlier voxels. Lower HD95 values indicate better boundary agreement between predicted and ground-truth annotations.

To complement segmentation accuracy with a measure of practical viability, two computational cost indicators were evaluated for every architecture: the mean inference time per test volume and the total number of trainable parameters. These indicators were analyzed jointly with the segmentation metrics to characterize the accuracy-efficiency trade-off of each model.

All experiments were executed on a single workstation equipped with an NVIDIA GeForce RTX~5090 GPU (24~GB VRAM), an Intel Core Ultra 9 275HX CPU, 64~GB of RAM, and a 1~TB NVMe SSD, running Microsoft Windows~11 Home 25H2 with the Windows Subsystem for Linux (WSL) and Ubuntu as the execution environment. Models were implemented in Python using PyTorch as the core deep learning framework and MONAI for medical-image-specific data loading, preprocessing transformations, reference architectures, loss functions, metric computation, and sliding-window inference. The \texttt{mamba\_ssm} library, together with dependencies such as \texttt{causal\_conv1d}, \texttt{einops}, and \texttt{transformers}, was used to run the state-space modules required by SegMamba and SegMambaV2. Auxiliary libraries included NumPy, SciPy, and Pandas for numerical, statistical, and tabular data handling, NiBabel for reading and writing NIfTI volumes, and Matplotlib for visualizing the results. Given the memory demands of volumetric MRI segmentation, all architectures were trained with spatial cropping and evaluated using sliding-window inference.

The code used to train and evaluate the models is available at \url{https://github.com/lunahernandez/unified-brats-benchmark.git} under the Apache License 2.0.

\section{Results}
\label{sec:results}

This section reports the outcome of the experimental framework described above. Results are organized around global and sub-region segmentation accuracy, spatial boundary error, the accuracy-efficiency trade-off, a qualitative three-dimensional assessment of the predicted masks, and comparisons with the top-performing approaches of the BraTS 2023 and BraTS 2024 challenges.

\subsection{Accuracy of Models}
Tables~\ref{tab:results_2023} and \ref{tab:results_2024} report the global Dice and IoU coefficients obtained on the test set for the BraTS 2023 and BraTS 2024 datasets, respectively. Metrics were averaged across the evaluated tumor sub-regions, excluding the background class. Across both clinical scenarios, the SSMs, SegMamba and SegMambaV2, consistently achieved the best overall performance.

\begin{table}[ht!]
    \centering
    \caption{Performance on the BraTS 2023 dataset: Dice, IoU, and HD95 metrics were averaged across the evaluated tumor sub-regions, excluding the background class. HD95 is reported in millimeters (lower is better).}
    \label{tab:results_2023}
    \begin{tabular}{lccc}
        \toprule
        \textbf{Model} & \textbf{Dice} & \textbf{IoU} & \textbf{HD95 (mm)} \\
        \midrule
        3D U-Net & 0.6737 & 0.5949 & 11.9 \\
        SegResNet & 0.6987 & 0.6246 & 11.1 \\
        Swin UNETR & 0.6607 & 0.5989 & 11.6 \\
        SegMamba & 0.7116 & 0.6437 & \textbf{9.2} \\
        SegMamba V2 & \textbf{0.7245} & \textbf{0.6574} & 9.5 \\
        \bottomrule
    \end{tabular}
\end{table}

\begin{table}[ht!]
    \centering
    \caption{Performance on the BraTS 2024 dataset: Dice, IoU, and HD95 metrics were averaged across the evaluated tumor sub-regions, excluding the background class. HD95 is reported in millimeters (lower is better).}
    \label{tab:results_2024}
    \begin{tabular}{lccc}
        \toprule
        \textbf{Model} & \textbf{Dice} & \textbf{IoU} & \textbf{HD95 (mm)} \\
        \midrule
        3D U-Net & 0.7142 & 0.6128 & 8.5 \\
        SegResNet & 0.7294 & 0.6362 & 7.9 \\
        Swin UNETR & 0.7391 & 0.6474 & 7.5 \\
        SegMamba & 0.7533 & 0.6649 & 7.3 \\
        SegMamba V2 & \textbf{0.7559} & \textbf{0.6678} & \textbf{7.1} \\
        \bottomrule
    \end{tabular}
\end{table}

For the intracranial meningioma task (BraTS 2023), SegMambaV2 led the performance with a Dice coefficient of 0.7245 and an IoU of 0.6574, closely followed by SegMamba (Dice = 0.7116, IoU = 0.6437). Interestingly, for this dataset, the convolutional SegResNet demonstrated strong intermediate results (Dice = 0.6987), whereas Swin UNETR yielded the lowest Dice score (0.6607), falling slightly behind 3D U-Net (0.6737).

In the post-treatment glioma dataset (BraTS 2024), SegMambaV2 maintained its superiority across all metrics, achieving the highest overlap (Dice = 0.7559, IoU = 0.6678), with SegMamba again providing similar metrics (Dice = 0.7533, IoU = 0.6649). In this scenario, the architectures followed a more conventional ranking: Swin UNETR recovered its expected performance tier to rank third (Dice = 0.7391), outperforming both SegResNet (Dice = 0.7294) and 3D U-Net, which recorded the lowest overlap (Dice = 0.7142). These results indicate that architectures incorporating state-space sequence modeling provide a robust and consistent advantage over purely convolutional and Transformer-based reference models, yielding superior overall segmentation accuracy across distinct brain tumor pathologies.

Regarding the Hausdorff Distance (HD95), which complements the overlap-based metrics by quantifying contour-level disagreement, in the BraTS 2023 dataset, the SSMs demonstrated the most precise boundary delineation. SegMamba obtained the best global error at 9.2\,mm, closely followed by SegMamba V2 (9.5\,mm). The residual convolutional architecture and the Transformer-based model presented higher spatial errors, with SegResNet achieving 11.1\,mm and Swin~UNETR 11.6\,mm. The purely convolutional baseline, 3D U-Net, registered the worst spatial error with 11.9\,mm.

This trend was similar in the more complex BraTS 2024 dataset. SegMamba V2 achieved the lowest global error (7.1\,mm), followed by SegMamba (7.3\,mm) and Swin~UNETR (7.5\,mm). SegResNet ranked fourth (7.9\,mm), while 3D U-Net once again exhibited the highest boundary disagreement (8.5\,mm).

Overall, the spatial-error analysis reinforces and extends the conclusions drawn from the Dice and IoU metrics. The Mamba-based models are not only top-performing in terms of global volumetric overlap, but they also achieve markedly superior boundary delineation across both clinical scenarios. This confirms that the architectural properties of state-space sequence modeling translate into a more robust and accurate reconstruction of tumor morphology, effectively minimizing the presence of spatially isolated outlier regions that heavily penalize boundary-sensitive metrics.

\subsection{Performance by Tumor Sub-region}
\label{subsec:subregion-results}

Global metrics conceal substantial variation across the annotated sub-regions, which differ significantly in prevalence and volume, as seen in Section~\ref{subsec:dataset}. Tables~\ref{tab:subregion_results_2023} and \ref{tab:subregion_results_2024} report the Dice score obtained by each architecture on the individual sub-regions for the BraTS 2023 and BraTS 2024 datasets, respectively.

\begin{table}[ht!]
    \centering
    \caption{Sub-region performance on the BraTS 2023 dataset: Dice coefficient obtained by each architecture on the evaluated tumor sub-regions. Best value per column in bold.}
    \label{tab:subregion_results_2023}
    \begin{tabular}{lccc}
        \toprule
        \textbf{Model} & \textbf{NETC} & \textbf{SNFH} & \textbf{ET} \\
        \midrule
        3D U-Net & 0.4131 & 0.7543 & 0.8537 \\
        SegResNet & 0.4409 & 0.7704 & 0.8847 \\
        Swin UNETR & 0.2953 & \textbf{0.7887} & 0.8982 \\
        SegMamba & 0.4527 & 0.7779 & 0.9040 \\
        SegMamba V2 & \textbf{0.4766} & 0.7875 & \textbf{0.9093} \\
        \bottomrule
    \end{tabular}
\end{table}

\begin{table}[ht!]
    \centering
    \caption{Sub-region performance on the BraTS 2024 dataset: Dice coefficient obtained by each architecture on the evaluated tumor sub-regions. Best value per column in bold.}
    \label{tab:subregion_results_2024}
    \begin{tabular}{lcccc}
        \toprule
        \textbf{Model} & \textbf{NETC} & \textbf{SNFH} & \textbf{ET} & \textbf{RC} \\
        \midrule
        3D U-Net & 0.5480 & 0.8532 & 0.7414 & 0.7140 \\
        SegResNet & 0.5613 & 0.8683 & 0.7565 & 0.7316 \\
        Swin UNETR & 0.5660 & 0.8755 & \textbf{0.7766} & 0.7381 \\
        SegMamba & \textbf{0.6085} & \textbf{0.8839} & 0.7608 & \textbf{0.7603} \\
        SegMamba V2 & 0.6072 & 0.8823 & 0.7750 & 0.7591 \\
        \bottomrule
    \end{tabular}
\end{table}

In BraTS 2023, the largest inter-model differences were observed for NETC, the least frequent (33.9\% of patient cases) and smallest-volume sub-region. Swin~UNETR struggled significantly with this class, obtaining a low Dice score of 0.2953, whereas SegMambaV2 achieved the best performance (0.4766). For SNFH, which was present in 53.6\% of cases, the results were more homogeneous, with Swin~UNETR narrowly taking the lead (0.7887) over SegMambaV2 (0.7875). The ET sub-region, present in 99.9\% of cases and with a comparatively large representative volume, yielded the highest scores overall, peaking with SegMambaV2 (0.9093) and SegMamba (0.9040).

The BraTS 2024 dataset introduces a fourth evaluation class: the resection cavity (RC). In this clinical scenario, SegMamba demonstrated exceptional robustness, obtaining the highest Dice scores for NETC (0.6085), SNFH (0.8839), and the RC class (0.7603). SegMambaV2 followed very closely across these three regions. However, the Transformer-based Swin~UNETR proved highly effective at delineating the enhancing tumor (ET) in this post-surgical environment, achieving the top score (0.7766), slightly ahead of SegMambaV2 (0.7750).

These results indicate a clear relationship between model characteristics and segmentation accuracy. The Mamba-based architectures maintained a dominant position across the majority of sub-regions in both clinical scenarios. Nevertheless, Swin~UNETR showed some strengths, particularly in SNFH (2023) and ET (2024), suggesting that its self-attention mechanisms remain highly competitive for specific anatomical delineations.

\subsection{Accuracy--Efficiency Trade-off}
\label{subsec:efficiency}

\begin{figure}[ht!]
    \centering
    \includegraphics[width=0.8\linewidth]{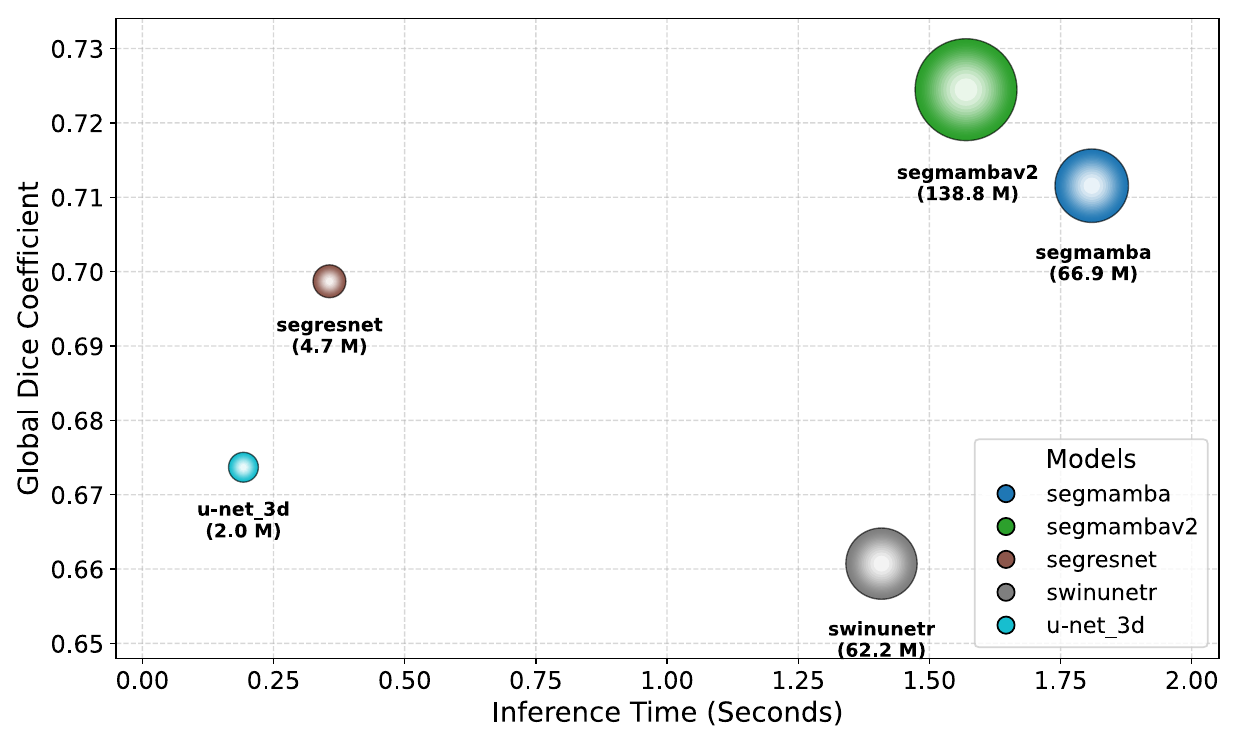}
    \caption{Accuracy--efficiency trade-off for BraTS 2023: Global Dice coefficient versus mean inference time per test volume for each architecture. Marker size encodes the number of trainable parameters for each model.}
    \label{fig:accuracy_vs_cost_2023}
\end{figure}

\begin{figure}[ht!]
    \centering
    \includegraphics[width=0.8\linewidth]{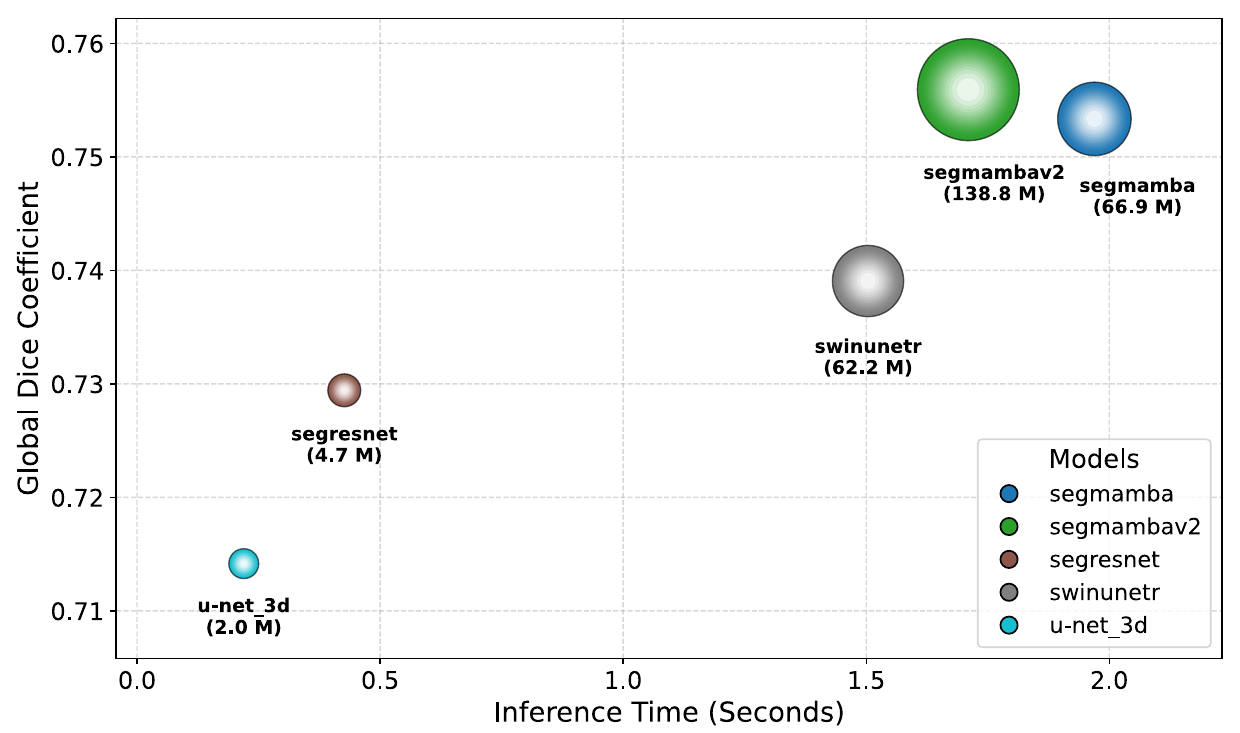}
    \caption{Accuracy--efficiency trade-off for BraTS 2024:  Global Dice coefficient versus mean inference time per test volume for each architecture. Marker size encodes the number of trainable parameters for each model.}
    \label{fig:accuracy_vs_cost_2024}
\end{figure}

Figures~\ref{fig:accuracy_vs_cost_2023} and~\ref{fig:accuracy_vs_cost_2024} show the trade-off between accuracy and efficiency for BraTS 2023 and 2024, respectively. These depict the relationship between global Dice, mean inference time per volume, and the total number of trainable parameters per model.

Across both clinical scenarios, 3D~U-Net proved to be the most computationally efficient architecture, featuring inference times of approximately 0.20--0.25\,s per patient and the lowest model complexity (2.0\,M parameters). However, this efficiency came at the cost of the lowest segmentation accuracy in the comparison. SegResNet offered a substantially better accuracy--efficiency balance among the convolutional models: it maintained sub-second inference (0.35--0.40\,s) and a moderate parameter count (4.7\,M) while achieving highly competitive Dice scores, making it an excellent candidate for hardware-constrained environments. 

Conversely, the Transformer-based Swin~UNETR required a significantly higher parameter count (62.2\,M) with inference times around 1.40--1.45\,s. Its segmentation accuracy varied notably across the clinical scenarios: while it yielded one of the lowest global overlaps in the 2023 dataset, it recovered to a highly competitive third position in the 2024 task. Nevertheless, since it did not achieve the highest global accuracy in either case, it ultimately presents a less favorable overall cost-to-performance ratio under the tested configuration compared to the state-space or lightweight convolutional alternatives.

The two state-space architectures occupied the most favorable region of the accuracy--efficiency space for high-performance segmentation. Both models achieved top-tier Dice scores with manageable computational requirements. Interestingly, SegMambaV2 proved to be faster in inference than the baseline SegMamba (approximately 1.55--1.70\,s versus 1.85--1.95\,s, respectively), despite having a substantially larger model complexity (138.8\,M versus 66.9\,M parameters). Taken together, these results indicate that the SegMamba family provides the best global trade-off for clinical scenarios where segmentation accuracy is paramount: SegMambaV2 offers a slight edge in inference speed and accuracy, whereas the original SegMamba achieves highly competitive results while requiring half the parameter count.

\subsection{Qualitative Assessment}
\label{subsec:qualitative}

\begin{figure}[htbp]
    \centering
    \includegraphics[width=0.8\linewidth]{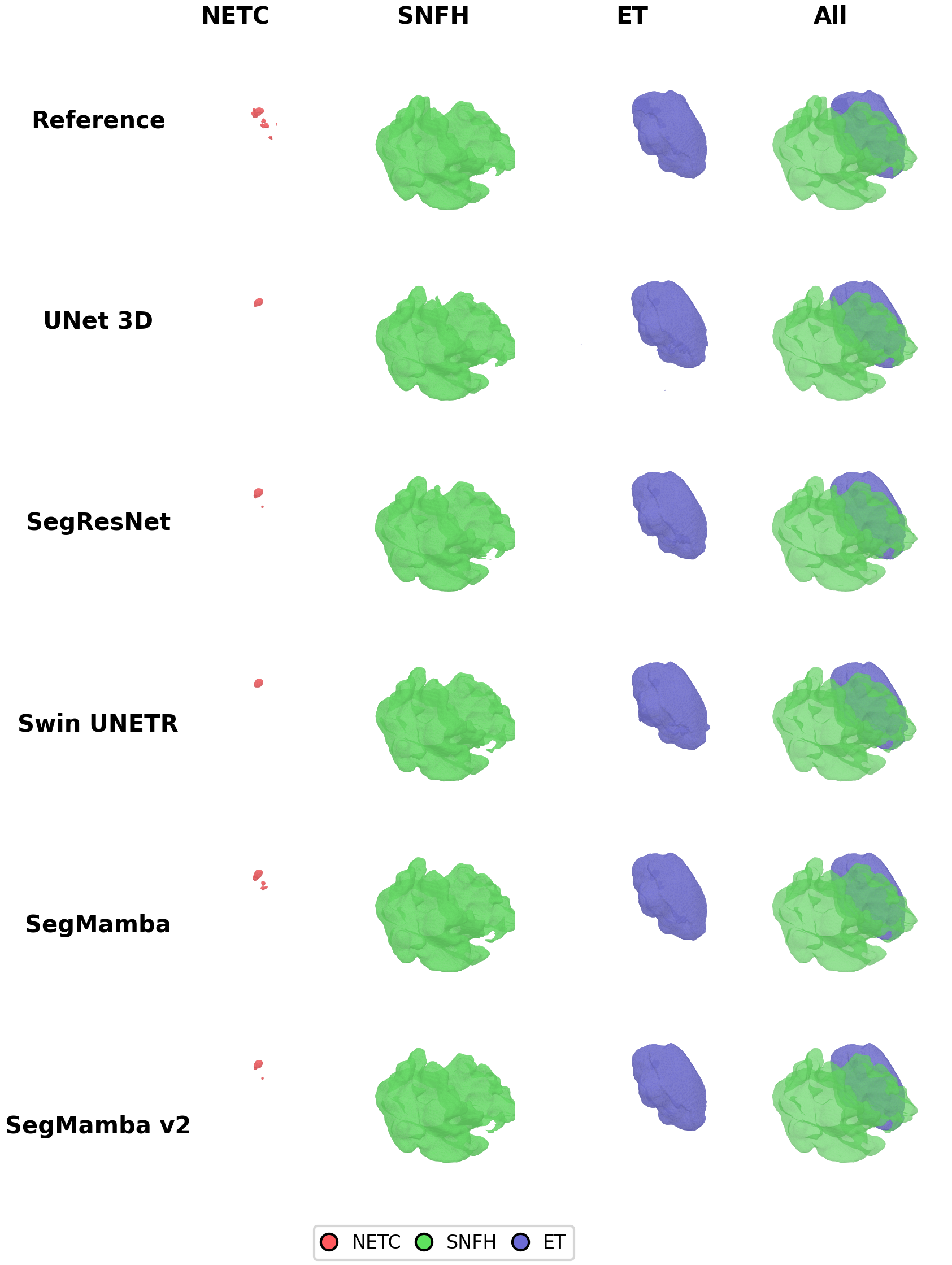}
    \caption{Three-dimensional comparison of the three tumor classes (NETC in red; SNFH in green; ET in blue) and their combination for the BraTS 2023 patient BraTS-MEN-00891-000. The ground-truth (reference) annotation is compared with predictions from the five models, showing each class in separate columns.}
    \label{fig:3d_classes_brats2023}
\end{figure}

\begin{figure}[htbp]
    \centering
    \includegraphics[width=0.5\linewidth]{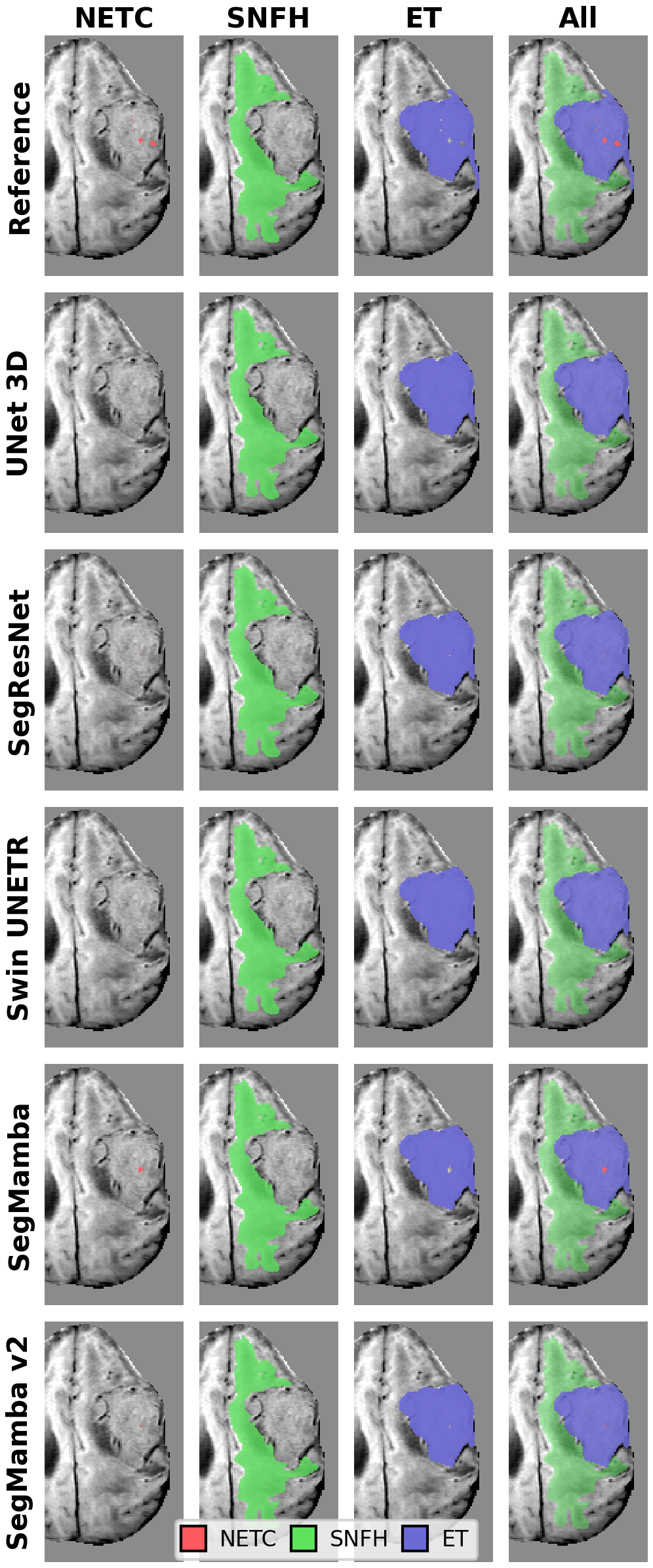}
    \caption{Two-dimensional axial comparison of predicted tumor classes superimposed on the T1c scan for the BraTS 2023 patient BraTS-MEN-00891-000. The ground-truth (reference) annotation is compared with predictions from the five models. The visualization displays the specific tumor classes (NETC in red; SNFH in green; ET in blue) in separate columns and combined.}
    \label{fig:2d_classes_brats2023}
\end{figure}

\begin{figure}[htbp]
    \centering
    \includegraphics[width=0.9\linewidth]{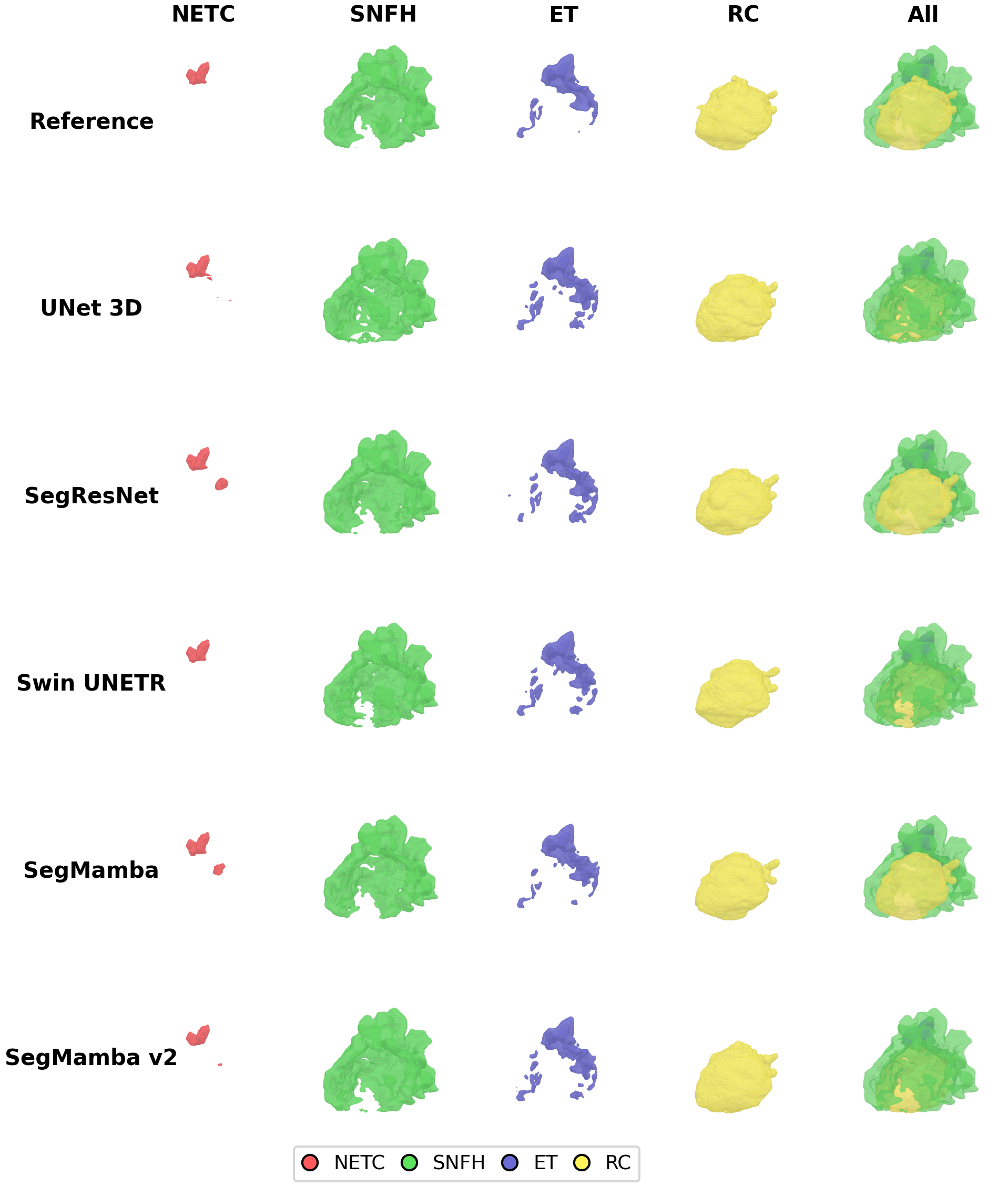}
    \caption{Three-dimensional comparison of the four tumor classes (NETC in red; SNFH in green; ET in blue; RC in yellow) and their combination for the BraTS 2024 patient BraTS-GLI-03063-100. The ground-truth (reference) annotation is compared with predictions from the five models, showing each class in separate columns.}
    \label{fig:3d_classes_brats2024}
\end{figure}

\begin{figure}[htbp]
    \centering
    \includegraphics[width=0.8\linewidth]{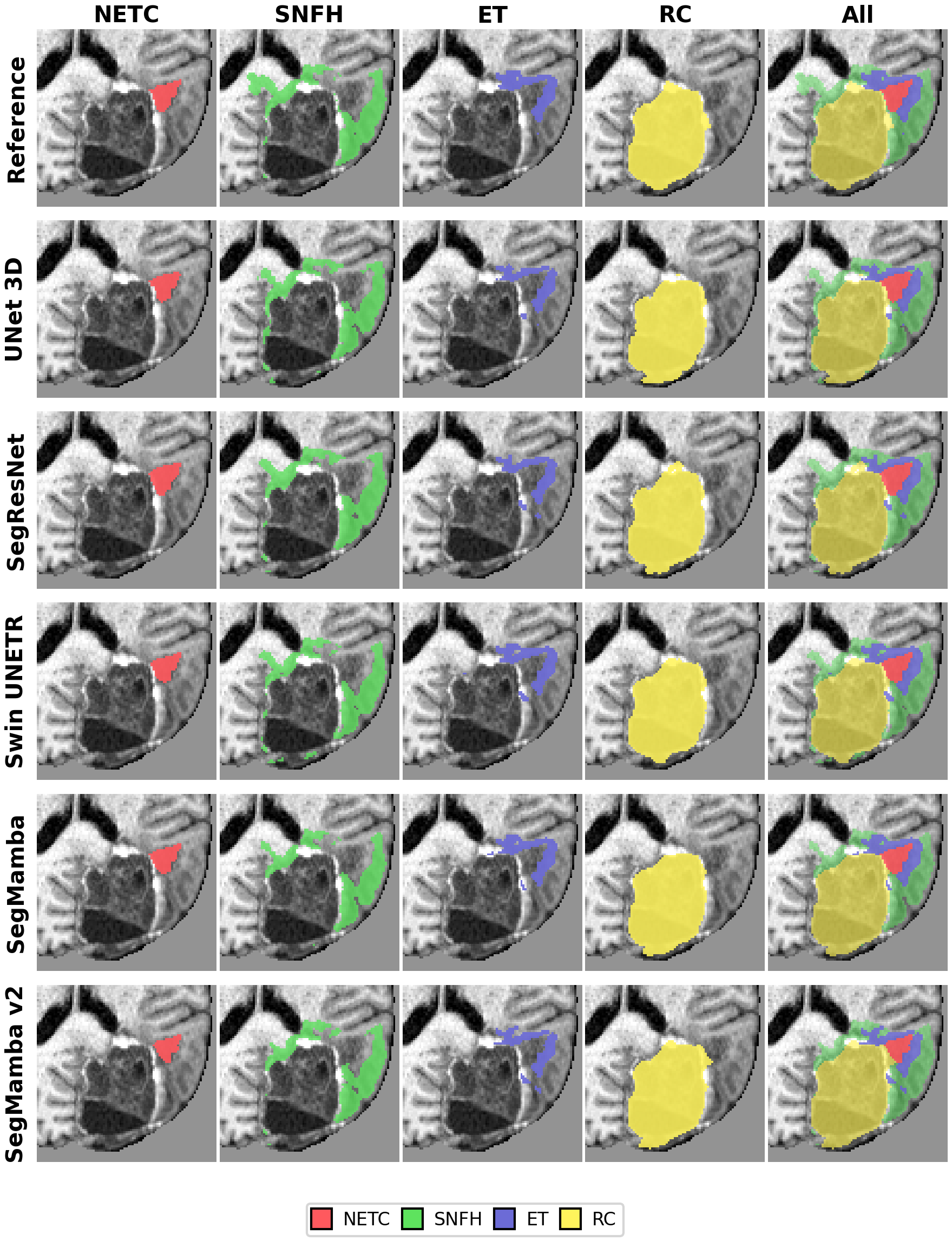}
    \caption{Two-dimensional axial comparison of predicted tumor classes superimposed on the T1c scan for the BraTS 2024 patient BraTS-GLI-03063-100. The ground-truth (reference) annotation is compared with predictions from the five models. The visualization displays the specific tumor classes (NETC in red; SNFH in green; ET in blue; RC in yellow) in separate columns and combined.}
    \label{fig:2d_classes_brats2024}
\end{figure}

To complement the quantitative evaluation, Figures~\ref{fig:3d_classes_brats2023}--\ref{fig:2d_classes_brats2023} and~\ref{fig:3d_classes_brats2024}--\ref{fig:2d_classes_brats2024} present three-dimensional and two-dimensional visualizations of the ground-truth and predicted segmentations, decomposed by tumor class, for patients from the BraTS 2023 and BraTS 2024 datasets, respectively.

For BraTS 2023 (Figures~\ref{fig:3d_classes_brats2023} and~\ref{fig:2d_classes_brats2023}), the SNFH and ET classes are reproduced closely by all five models, with predicted contours that closely resemble the ground-truth annotation in both the volumetric renderings and the axial slice. The NETC class is noticeably harder to segment: its scattered, multifocal morphology in the reference mask results in small, isolated components that are difficult for the models to recover consistently, making NETC the class with the greatest disagreement among predictions for this patient.

For BraTS 2024 (Figures~\ref{fig:3d_classes_brats2024} and~\ref{fig:2d_classes_brats2024}), differences between models are more pronounced, particularly for the SNFH and NETC classes, where false positives are readily identifiable in both the individual class views and the combined ("All") visualization. Differences in the ET and RC classes are comparatively more subtle: ET is fragmented into numerous small, scattered regions across all models, making inter-model discrepancies harder to discern visually, whereas RC's more regular and compact morphology is reconstructed consistently across models.

\subsection{Comparison with the State-of-the-Art}
\label{subsec:sota-results}

To contextualize the performance of the evaluated architectures, the models were compared against the top-ranked approaches from their respective challenges. For BraTS 2023, we compared our results against Auto3DSeg~\cite{he2023auto3dseg} and, for BraTS 2024, against Ferreira et al.~\cite{ferreira2024improved}, which employed an ensemble of nnU-Net and MedNeXt models augmented with synthetic data generation.

Nevertheless, three methodological differences make this comparison indicative rather than conclusive. First, these state-of-the-art references utilized ensembling strategies and hundreds of training epochs, whereas the architectures in this study were trained under a constrained, single-model protocol. Second, there is a discrepancy in data availability, since these models were trained on the entire provided training dataset and evaluated on the held-out validation set. In contrast, the present study evaluated the models using a local train-validation-test split, effectively reducing the volume of data available for training. Finally, the evaluation metrics differ fundamentally in their calculation. The challenge portal reports \textit{lesion-wise} metrics, which evaluate each connected component as an independent object. This approach penalizes spatial metrics, such as HD95, if a model completely misses a distinct lesion or hallucinates a small, distant false positive. In contrast, the models in this study are evaluated using standard global metrics, often referred to as \textit{voxel-wise}, which measure the overall volumetric agreement. Because standard HD95 computes distances across the entire predicted volume globally, it is more tolerant to small isolated errors provided the main tumor bulk is accurately delineated. Consequently, numeric comparisons, particularly for spatial boundary errors, must be interpreted with this methodological gap in mind.

Table~\ref{tab:sota-comparison-2023} summarizes the comparison for BraTS 2023. The state-space architectures (SegMamba and SegMambaV2) achieved highly competitive Dice scores that rival and even marginally exceed the Auto3DSeg ensemble reference (e.g., SegMambaV2 reached a Dice of 0.9093 on ET and 0.9052 on TC, compared to the ensemble's 0.8985 and 0.9035, respectively). The most dramatic numeric difference is observed in the HD95 metric, where the local models exhibit spatial errors under 10\,mm, compared to the 21--31\,mm range of Auto3DSeg. This discrepancy is primarily due to the differences in the aforementioned evaluation methodology. 

\begin{table}[ht!]
\centering
\caption{Comparison with the state-of-the-art on the composite regions (ET, TC, WT) for BraTS 2023. Best value per column among the models evaluated in this study is in bold.}
\label{tab:sota-comparison-2023}
\begin{tabular}{p{3cm}cccccc}
\toprule
\textbf{Model} & \textbf{Dice ET} & \textbf{Dice TC} & \textbf{Dice WT} & \textbf{HD95 ET} & \textbf{HD95 TC} & \textbf{HD95 WT} \\
\midrule
Auto3DSeg~\cite{he2023auto3dseg}   & 0.8985 & 0.9035 & 0.8709 & 23.86 & 21.82 & 31.39 \\
\midrule
3D~U-Net~\cite{cciccek20163d}      & 0.8537 & 0.8484 & 0.8358 & 13.14 & 13.84 & 14.51 \\
SegResNet~\cite{SegResNet}         & 0.8847 & 0.8805 & 0.8663 & 13.54 & 14.12 & 15.13 \\
Swin~UNETR~\cite{Swin_UNETR}       & 0.8982 & 0.8935 & 0.8761 & 13.28 & 13.68 & 14.16 \\
SegMamba~\cite{xing2024segmamba}   & 0.9040 & 0.8988 & 0.8896 & \textbf{8.70} & \textbf{9.22} & \textbf{9.17} \\
SegMambaV2~\cite{xing2025segmamba} & \textbf{0.9093} & \textbf{0.9052} & \textbf{0.8962} & 9.29 & 9.74 & 9.48 \\
\bottomrule
\end{tabular}
\end{table}

Table~\ref{tab:sota-comparison-2024} extends this analysis to BraTS 2024. Against Ferreira et al.'s ensemble method, the evaluated single-model architectures demonstrated remarkable performance despite training on fewer cases. Swin~UNETR achieved the highest Dice among our tested architectures for ET (0.7766) and TC (0.7656), close to the SOTA model. For the WT region, SegMamba (0.8916) matched the reference model (0.8938). The global HD95 values for all evaluated models (between 5.70\,mm and 8.24\,mm) are significantly lower than the lesion-wise spatial errors reported by Ferreira et al. (17.95--38.19\,mm), although this discrepancy is in part due to the differences in the evaluation methodology.

\begin{table}[ht!]
\centering
\caption{Comparison with the state-of-the-art on the composite regions (ET, TC, WT) for BraTS 2024. Best value per column among the models trained in this study is in bold.}
\label{tab:sota-comparison-2024}
\begin{tabular}{p{3cm}cccccc}
\toprule
\textbf{Model} & \textbf{Dice ET} & \textbf{Dice TC} & \textbf{Dice WT} & \textbf{HD95 ET} & \textbf{HD95 TC} & \textbf{HD95 WT} \\
\midrule
Ferreira et al. (Ensemble)~\cite{ferreira2024improved} & 0.7900 & 0.7874 & 0.8938 & 35.63 & 38.19 & 17.95 \\
\midrule
3D~U-Net~\cite{cciccek20163d}      & 0.7414 & 0.7332 & 0.8662 & 7.54 & 8.24 & 6.93 \\
SegResNet~\cite{SegResNet}         & 0.7565 & 0.7464 & 0.8790 & 6.60 & 7.18 & 6.56 \\
Swin~UNETR~\cite{Swin_UNETR}       & \textbf{0.7766} & \textbf{0.7656} & 0.8849 & 6.20 & 6.82 & 6.00 \\
SegMamba~\cite{xing2024segmamba}   & 0.7607 & 0.7483 & \textbf{0.8916} & 6.66 & 7.33 & 5.78 \\
SegMambaV2~\cite{xing2025segmamba} & 0.7750 & 0.7626 & 0.8904 & \textbf{6.07} & \textbf{6.68} & \textbf{5.70} \\
\bottomrule
\end{tabular}
\end{table}

These patterns suggest that despite the handicap of a reduced training set and the absence of complex ensembling techniques or synthetic data generation, the evaluated architectures---particularly the SegMamba family and Swin~UNETR---are capable of producing highly accurate and competitive segmentations.

\section{Conclusion}
\label{sec:conclusion}

This work presented a unified and reproducible benchmark for evaluating representative CNN-, Transformer-, and SSM-based architectures for 3D brain tumor segmentation. Unlike many previous studies that evaluate models under heterogeneous experimental settings or on a single dataset, all architectures were trained and assessed using identical preprocessing, optimization, inference, and evaluation protocols across two complementary BraTS benchmarks representing distinct clinical scenarios. The BraTS 2023 meningioma dataset primarily evaluates the ability of segmentation models to accurately delineate relatively compact tumors with well-defined boundaries, and the BraTS 2024 post-treatment glioma dataset introduces substantially greater anatomical complexity due to heterogeneous tissue appearance, treatment-induced alterations, and ambiguous lesion boundaries.

The experimental results demonstrated that all evaluated architectures were capable of producing accurate brain tumor segmentations, although important differences emerged in terms of segmentation accuracy, boundary delineation, computational efficiency, and robustness. SSM architectures, SegMamba and SegMambaV2, consistently achieved the best global performance across both datasets, outperforming the CNN- and Transformer-based baselines in most evaluation metrics. Nevertheless, the results also showed that Transformer-based models remained highly competitive for specific tumor sub-regions, being especially beneficial for complex anatomical structures where spatial relationships extend beyond the receptive field of conventional convolutions. Convolutional architectures, on the other hand, continue to offer an attractive balance between accuracy and computational cost.

Despite the comprehensive evaluation performed in this study, it presents several limitations: The evaluation was restricted to two datasets and a representative selection of segmentation architectures, while alternative model variants, optimization strategies, or additional clinical datasets could produce different performance rankings; Furthermore, the benchmark focused exclusively on supervised learning using annotated data, without considering other paradigms such as self-supervised pretraining or foundation models.

Future work will extend the proposed benchmark by incorporating additional datasets, recently proposed foundation models, uncertainty estimation techniques, and explainable artificial intelligence methods. These extensions will contribute to a broader understanding of the capabilities and limitations of next-generation medical image segmentation algorithms and facilitate their translation into real clinical practice.

\bibliographystyle{unsrt}  
\bibliography{references}  

\end{document}